\pdfoutput=1

\documentclass[11pt]{article}
\usepackage[table]{xcolor}
\usepackage{pgf}
\definecolor{colorLow}{rgb}{1,1,0.6} 
\definecolor{colorHigh}{rgb}{1,0,0}  
\usepackage{multirow}
\newcommand{\HeatCell}[1]{
    \pgfmathsetmacro{\TempPercentColor}{((#1-19)/(60-19)*100)}
    \pgfmathsetmacro{\PercentColor}{max(0, \TempPercentColor)}
    \pgfmathsetmacro{\PercentColor}{min(100, \PercentColor)}
    \xdef\PercentColor{\PercentColor}
    \cellcolor{cyan!\PercentColor!white}#1
}

\newcommand{\HeatCellGPT}[1]{
    \pgfmathsetmacro{\TempPercentColor}{((#1-14)/(50-14)*100)}
    \pgfmathsetmacro{\PercentColor}{max(0, \TempPercentColor)}
    \pgfmathsetmacro{\PercentColor}{min(100, \PercentColor)}
    \xdef\PercentColor{\PercentColor}
    \cellcolor{cyan!\PercentColor!white}#1
}

\newcommand{\HeatCellLosses}[1]{
    \pgfmathsetmacro{\TempPercentColor}{((#1-10)/(80-10)*100)}
    \pgfmathsetmacro{\PercentColor}{max(0, \TempPercentColor)}
    \pgfmathsetmacro{\PercentColor}{min(100, \PercentColor)}
    \xdef\PercentColor{\PercentColor}
    \cellcolor{cyan!\PercentColor!white}#1
}

\newcommand{\HeatCellRuleBERTMain}[1]{
    \pgfmathsetmacro{\TempPercentColor}{(pow(2, (#1-70)/10) - 1)/(pow(2, (86-70)/10) - 1)*100}
    \pgfmathsetmacro{\PercentColor}{max(0, \TempPercentColor)}
    \pgfmathsetmacro{\PercentColor}{min(100, \PercentColor)}
    \xdef\PercentColor{\PercentColor}
    \cellcolor{cyan!\PercentColor!white}#1
}

\newcommand{\HeatCellRuleBERT}[1]{
    \pgfmathsetmacro{\TempPercentColor}{(pow(2, (#1-63)/10) - 1)/(pow(2, (87-63)/10) - 1)*100}
    \pgfmathsetmacro{\PercentColor}{max(0, \TempPercentColor)}
    \pgfmathsetmacro{\PercentColor}{min(100, \PercentColor)}
    \xdef\PercentColor{\PercentColor}
    \cellcolor{cyan!\PercentColor!white}#1
}

\usepackage[final]{acl}

\usepackage{times}
\usepackage{latexsym}

\usepackage[T1]{fontenc}

\usepackage[utf8]{inputenc}

\usepackage{microtype}

\usepackage{fontawesome5}
\usepackage{hyperref}
\usepackage{graphicx}
\usepackage{booktabs}
\graphicspath{ {./images/} }
\usepackage{changepage}
\usepackage{hyperref}
\usepackage{url}
\usepackage{subcaption}
\usepackage[]{graphicx} 
\usepackage{algorithm}
\usepackage{algpseudocode}
\usepackage{amsmath}
\usepackage{amssymb}
\algtext*{EndWhile}
\algtext*{EndIf}
\algtext*{EndFor}

\title{Teaching Probabilistic Logical Reasoning to Transformers }

\author{Aliakbar Nafar \\
Michigan State University \\
  \texttt{nafarali@msu.edu} \\\And
  Kristen Brent Venable \\
  Florida Institute for Human \\
  and Machine Cognition \\
  \texttt{bvenable@ihmc.org} \\\And
  Parisa Kordjamshidi \\
  Michigan State University \\
  \texttt{kordjams@msu.edu} \\}

\begin{document}

\maketitle
\begin{abstract}

In this paper, we evaluate the capability of transformer-based language models in making inferences over uncertain text that includes uncertain rules of reasoning. We cover both Pre-trained Language Models (PLMs) and generative Large Language Models (LLMs). Our evaluation results show that both generations of language models struggle with reasoning over uncertain text. We propose a novel end-to-end fine-tuning approach, Probabilistic Constraint Training (PCT), that utilizes probabilistic logical rules as constraints in the fine-tuning phase without relying on these rules in the inference stage. To assess the effectiveness of PCT, we utilize the related corpora and, additionally, create a new and more challenging benchmark that, unlike the previous ones, uses instance-specific rules. Our study demonstrates that PCT improves the transformer-based language model's intrinsic reasoning and makes their probabilistic logical reasoning process more explicit and explainable. Furthermore, PCT equips these models to effectively handle novel situations, including higher reasoning depth, new domains, and complex probabilistic structures.

\end{abstract}

\section{Introduction}
\label{sec:introduction}

Language models have demonstrated high performance across a wide range of Natural Language Processing (NLP) tasks~\cite{DBLP:journals/corr/abs-1907-11692} which in the case of Large Language Models holds even in zero-shot setting~\cite{chen-2023-large}. However, they struggle to reason over uncertain text involving logical probabilistic rules~\cite{saeed-etal-2021-rulebert,jin2023CLADDER}. This is confirmed by the reported poor results in arithmetic reasoning when using transformers~\cite{mishra-etal-2022-numglue} which is required for probabilistic logical reasoning. 
Additionally, logical probabilistic inference requires coherent step-by-step reasoning. 
However, PLMs' evaluation of various question-answering (QA) benchmarks shows they produce contradictory results that violate the expected steps of reasoning, such as following transitivity or symmetry rules~\cite{asai-hajishirzi-2020-logic}. 
This has led to the development of hybrid approaches, where reasoning tasks are outsourced to Neuro-Symbolic engines, bypassing the need for reasoning by transformers~\cite{zhang-etal-2023-improved}.
To overcome these limitations, we embed probabilistic reasoning into transformers by imposing the rules of logical probabilistic reasoning as constraints during their training phase.

\begin{table*}[!t]
\begin{small}

  \centering
  \def\arraystretch{1.5}%
  \begin{tabular}{|p{7.5cm}|p{7.5cm}|}
  \hline
  \hline
     RuleBERT & RuleTaker-pro \\
     \hline
     \hline
    (Fact 1) David is a cousin of Ann.
    
(Fact 2) Mike is a child of Ann.

(Rule 1, 0.90) If A is a spouse of B and C is a child of B,
 then C is a child of A. 
 
(Rule 2, 0.15) If A is a cousin of B, then A is a spouse of B. 
 & (Fact 1) Dave is big.
 
(Fact 2) Erin is sad.

(Rule 1) Usually, If someone is big then they are green.

(Rule 2) Normally, If someone is green then they are round.

(Rule 3) Seldom, If someone is sad then they are round.  \\
    \hline
    (Query) Mike is a child of David.
 &  (Query) Dave is round. \\
    \hline
    
    \multicolumn{2}{|c|}{Required Steps of Reasoning to Answer}\\
    \hline
Fact 1 (1.00) \& Rule 2 (0.15) $\implies$ 
    
Fact 3: David is a spouse of Ann. (0.15) (Inferred)

Fact 3 (0.15) \& Fact 2 (1.00) \& Rule 1 (0.90) $\implies$ 

Fact 4: Mike is a child of David. (0.135) (Inferred) 
 
\textbf{Answer: 0.135} & Fact 1 (1.00) \& Rule 1 (0.90) $\implies$ 
 
 Fact 3: Dave is green. (0.90) (Inferred) 

Fact 3 (0.90) \& Rule 2 (0.80) $\implies$ 

Fact 4: Dave is round. (0.72) (Inferred)

\textbf{Answer: 0.72} \\
 \hline
 \multicolumn{2}{|c|}{Approach: Converting Probabilistic Reasoning Steps to Equality Constraints}\\
    \hline

Constraint 1: P( Fact 1 ) * 0.15 = P( Fact 3 )
    
Constraint 2: P( Fact 3 ) * P( Fact 2 ) * 0.90 = P( Fact 4 )& 
 
Constraint 1: P( Fact 1 ) * 0.90 = P( Fact 3 )
 
Constraint 2: P( Fact 3 ) * 0.80 = P( Fact 4 )\\
    \hline

  \end{tabular}
  \caption{Left column: an example from RuleBERT with two facts and two rules. Right column: an example from RuleTaker-pro with two facts and three rules. The reasoning steps required to infer the Query and the constraints applied in these steps are shown in the bottom rows.}
  \label{table:examples}

\end{small}
  
\end{table*}
There are only a few research efforts dealing with uncertainty in text. Understanding logical and uncertain rules in natural language form has been investigated in recent research on question answering~\cite{DBLP:journals/corr/abs-2002-05867,saeed-etal-2021-rulebert}, and there have been several attempts to teach transformers how to follow these rules~\cite{asai-hajishirzi-2020-logic,faghihi2023gluecons}. 
While incorporating hard logical rules is undoubtedly important and is still being investigated, in the real world, most of the external knowledge and rules involve uncertainty. For example, only a small fraction of the logical rules in DBpedia can be deemed certain~\cite{saeed-etal-2021-rulebert}. Inference over text that includes uncertainty concerning facts, relations, and rules is required in many natural language comprehension tasks. For example, scientific content often utilizes hedges to express the measure of certainty in factual statements~\cite{DBLP:journals/corr/abs-2109-14776,PMID28406600}. 

A related but different challenge is the explainability of the solutions provided by transformer-based language models. Without the capability of providing the underlying components and steps necessary to answer a question, a Language Model's reasoning remains inexplicable even when it accurately answers a question~\cite{clark-etal-2019-bert}. In this paper, we propose a  method that forces the transformer to follow coherent reasoning steps to answer the final question, as shown in Table~\ref{table:examples}, yielding a more explainable model. This feature is an inherent property and a byproduct of our usage of probabilistic logical constraints and Neuro-Symbolic modeling in our approach.


In this paper, to deal with reasoning over uncertain text, we look into a problem setting that involves calculating the probability of a given hypothesis (Query) based on a provided context that includes \textit{linguistic expression} of probabilistic logical rules and facts. The underlying reasoning is probabilistic logical inference. We utilize two QA datasets: RuleBERT~\cite{saeed-etal-2021-rulebert} and our newly developed RuleTaker-pro, created to include context-specific rules. Table~\ref{table:examples} shows examples of our datasets and their required reasoning steps to answer the Query. We convert the reasoning steps to equality constraints (shown in the Approach section of Table~\ref{table:examples}) and impose these constraints to ensure consistency of the outputs with the rules during the training of PLMs but not inference. Despite the simplicity of the reasoning patterns in our approach, we will show the transferability of learning to more complex structures. In summary, our contributions are as follows:

\noindent 1) We propose a new approach, Probabilistic Constraint Training (PCT), that explicitly imposes probabilistic reasoning rules during PLM fine-tuning. This approach provides an effective level of abstraction to the models to generalize and transfer reasoning under uncertainty to new domains and to more complex depths of reasoning. 2) We develop a novel evaluation benchmark for probabilistic reasoning over text with context-specific uncertain rules whose probabilities can not be captured from the training data and must be extracted from the text.\footnote{The code and dataset are available at \href{https://github.com/HLR/Probabilistic_Logical_Reasoning}{\faGithub}.} 3) We conduct thorough experiments comparing our constraint-based fine-tuning approach with LLMs and show the superiority of our technique.

\section{Related work}
\label{sec:relatedwork}
Previous works mostly looked into the integration of crisp logic~\cite{saha-etal-2020-prover,tafjord-etal-2021-proofwriter}. The earlier work on QA with probabilistic rules in the text is RuleBERT~\cite{saeed-etal-2021-rulebert}, which serves as the baseline for our comparative study. While RuleBERT pioneers this field and introduces Weighted binary cross-entropy loss to incorporate probabilistic learning in transformers, it lacks a mechanism to follow the probabilistic reasoning steps explicitly. Additionally, our experiments revealed that the rules in textual form in this dataset are not properly utilized by the models (see Section~\ref{sec:exp1}), which prompted us to introduce RuleTaker-pro with context-specific rules.

\noindent\textbf{Reasoning Steps.} Explicit elucidation of reasoning steps in QA models has been central in recent literature. \citeauthor{saha-etal-2020-prover} improve PLMs' reasoning by mapping their output to an inference graph, necessitating the model to learn its nodes and edges. While \citeauthor{tafjord-etal-2021-proofwriter} utilize T5 to create an inference path, this and similar studies have focused on using non-probabilistic logical rules. In several other related works, the reasoning for QA is approached by generating an output that follows a predefined formal language for theorem proving given the logical rules, which is a very different approach from ours~\cite{NEURIPS2020_d2a27e83,DBLP:journals/corr/abs-2009-03393,tafjord-etal-2021-proofwriter}. \citeauthor{wu-etal-2023-chain} introduce reasoning in LLMs by generating intermediate reasoning steps as an extra output. However, we enable PLMs to incorporate this reasoning in training with no additional output.



\noindent\textbf{Constraints.} Our approach's primary contribution is incorporating probabilistic constraints in the loss function. While various studies incorporate logical constraints into the loss function~\cite{NEURIPS2019_cf708fc1,li-etal-2019-logic,asai-hajishirzi-2020-logic,ribeiro-etal-2019-red,faghihi2023gluecons,ijcai2020p382}, no work has explored the application of probabilistic constraints in this context to date. 


\noindent\textbf{Neuro-Symbolic Methods.} Central to our approach is the implementation of an end-to-end model, ensuring the transferability of our model to various domains without the need to modify the model's architecture or decision processes. In contrast, numerous studies in this field rely on a \textbf{pipeline} approach, often incorporating a \textbf{Neuro-Symbolic engine}. \citeauthor{zhang-etal-2023-improved} proposes a framework in which Transformers extract the factual knowledge in the text. Consequently, a symbolic engine conducts the reasoning inference. More general approaches use deep neural methods to process the input and use either an existing engine~\cite{NEURIPS2018_dc5d637e} such as Problog~\cite{10.5555/1625275.1625673} or define a language to create a logical structure as an inference engine~\cite{10.1145/3591280}.

\section{Background}

\subsection{Problem Definition} We focus on the challenge of performing probabilistic logical reasoning within a QA task where a set of facts $F$, a set of rules $R$, and a hypothesis $h$ are provided in a \textbf{textual} context. While these rules, facts, and hypothesis are provided \textbf{only} in their \textbf{textual form} as a part of the input to the task, we have their formal information as a part of the metadata. For example, fact $Big(Dave)$ and the rule $Spouse(A, B)\ \&\ Child(C, B) \rightarrow Child(C, A)$ would be given as input in forms: ``Dave is big.'', and  ``If A is a spouse of B and C is a child of B, then C is a child of A.'', respectively. The facts and hypothesis consist of factoids that define properties for an entity ``Has\_Property(Entity)'' or relations between two entities ``Relation(Entity1, Entity2)''. The rules have the form $(p_1,p_2,...,p_n) \to q, Pr$, where $p_i$ represents a premise fact, $q$ is a new inferred fact, and $Pr$ is the probability of the rule. In RuleBERT rules, $Pr$ is not directly mentioned in the rule's text and must be learned from the data (or extracted from the metadata to be used in the loss during training), while in RuleTaker-pro, $Pr$ is mentioned in the form of adverbs of uncertainty. $q$'s probability is computed as the rule probability multiplied by the premise facts probabilities. If the premise facts are mentioned in the context, they would be certain and have a probability of $1.00$; otherwise, if they are inferred facts, their probability is derived. The objective is to utilize $F$ and $R$ to infer a probability between 0 and 1 as our task output, which indicates the probability of a given hypothesis $h$. For example, h=``Sara and John are cousins'' obtains a probability of 0.20 by the model. 

\subsection{Base Model} The backbone of our model is RoBERTa Large, supplemented by two linear layers and a sigmoid activation function applied to its classifier token (CLS). The model takes the textual representations of facts and rules (context) and hypothesis as input formatted as [CLS] text(R)+text(F) [SEP] text(h) [SEP]. Subsequently, it predicts a probability for the given hypothesis.

The LLM models that we use as baselines for comparisons are GPT3.5 and GPT4~\cite{NEURIPS2020_1457c0d6}. Due to the high cost of fine-tuning LLMs, we limit our experiments to zero-shot and few-shot. Input comprises a task explanation, text(R)+text(F), and text(h). The explanation instructs the model about the objective and output format, either ``True'',``False'' (corresponding to a probability greater or less than 0.5), or the hypothesis probability (between 0.0 and 1.0).

\subsection{Deep Learning with Logical Constraints} Among the research focused on constraint integration within neural models, we opt for the class of methods that incorporate constraint violation in the loss function during training without altering the model's architecture~\cite{NEURIPS2019_cf708fc1,li-etal-2019-logic,faghihi2023gluecons}. In general, to employ the logical and symbolic constraints in deep models, they must be converted into soft logic for the sake of differentiability. Usually, three main approaches are used for this conversion: Product, Gödel, and  Łukasiewicz \cite{li-etal-2019-logic}. For instance, the logical rule, $(p_1,p_2,...,p_n) \to q$, using the Product surrogate, is written as follows,

\begin{small}
\begin{equation}
\setlength{\abovedisplayskip}{1pt}
\setlength{\belowdisplayskip}{4pt}
\label{eqn:constraint2}
min(1,P(q)/ [ P(p_1)*P(p_2)*...*P(p_n) ]),
\end{equation}
\end{small}
\noindent where $P(p_i)$ is the probability of the fact $p_i$. 
We can express the enforcement of this implication's truth as follows,

\begin{small}
\begin{equation}
\setlength{\abovedisplayskip}{1pt}
\setlength{\belowdisplayskip}{4pt}
\label{eqn:constraint2expresed}
\lvert 1-min(1,P(q)/ [P(p_1)*P(p_2)*...*P(p_n) ] )\rvert=0,
\end{equation}
\end{small}

\noindent where |.| denotes the absolute value. These methods of constraint conversion are defined for logical constraints and \textbf{do not directly apply to probabilistic reasoning rules}, which is why we will introduce a novel method of constraint integration for our goal of enforcing probabilistic reasoning.

\section{Training with Probabilistic Constraints}
We aim to develop a model capable of following probabilistic reasoning steps to infer the probability of a given hypothesis. These reasoning steps for the examples in Table~\ref{table:examples} are outlined in the \textit{Required Steps of Reasoning to Answer} row. In each step, a combination of facts and a rule results in a new \textit{intermediate inferred fact} until the final hypothesis is inferred. These steps are formulated as constraints, and our proposed model is trained to adhere to them by incorporating them into the loss function. The \textit{Approach} row of Table~\ref{table:examples} shows examples of the reasoning steps' conversion into constraints in which the probabilities assigned to facts must follow the rule definition. For instance, if Fact 1 and Rule 1 result in a new fact, Fact 1's probability (P(Fact 1) multiplied by Rule 1's probability must be equal to the inferred fact's probability (P(inferred Fact)). In the upcoming subsections, we will explain the process of formulating and utilizing constraints.

\subsection{Constraint Integration} 
\label{sec:cosntraint_conversion}

We formulate the probabilistic reasoning as obeying a set of constraints derived from probabilistic inference calculations, based on an assumed probabilistic network. We distinguish between \textit{Simple} and \textit{Complex} probabilistic reasoning patterns based on their underlying inference network. A probabilistic reasoning pattern is \textit{Simple} if any deducible fact can be drawn from it via only a single reasoning path. The examples provided in Table~\ref{table:examples} are \textit{Simple} because ``Dave is round.'' can be inferred only from Rule 2 and Fact 3 and Fact 3 can only be inferred from Fact 1 and  Rule 1. On the other hand, a \textit{Complex} reasoning encompasses at least one fact that can be deduced from two or more different rules (reasoning paths). By altering the second fact from ``Erin is sad'' to ``Dave is sad'', we create a \textit{Complex} example because it enables inference of ``Dave is round'' from Fact 2 and Rule 3 as well. Our focus lies primarily on formulating the simple version of probabilistic reasoning for defining constraints. The \textit{Complex} examples are still incorporated in our datasets and used during training and testing. Later, we investigate how our proposed model can also generalize over the \textit{Complex} probabilistic networks.

Given a \textit{Simple} network, our model executes probabilistic inference for the rule $(p_1,p_2,...,p_n) \to q, Pr$ by multiplying the probability of premise facts by the probability of the rules to obtain the probability of the inferred fact. Formally, the model should fulfill the constraint,

\begin{small}
\begin{equation}
\setlength{\abovedisplayskip}{1pt}
\setlength{\belowdisplayskip}{4pt}
\label{eqn:constraint1}
 \lvert P(q)-P(p_1)*P(p_2)*...*P(p_n)*Pr\rvert=0.
\end{equation}
\end{small}
\noindent Our unique definition of constraint constitutes the key novelty of our approach (see Table~\ref{table:examples} for Examples of constraints). To satisfy this constraint, the left side of the above equation should approach zero. Note that while this constraint guarantees adherence to the probabilistic rules, it might not ensure the best results on the end task accuracy, and this remains subject to experimentation.

\subsection{Training and Inference}

\noindent \textbf{Training} To generate the constraints for each dataset example, we use the chains of probabilistic reasoning that include the paths of inference for every inferable fact  (available in the dataset metadata; see section~\ref{datasetcreation}). Examples of these constraints can be found at the bottom of Table~\ref{table:examples}. We denote the violation from each constraint as $C_i$, a scalar value that ranges from 0 to 1, that is, the left-hand side of Equation~\ref{eqn:constraint1}. Our training objective centers on minimizing the violation of these constraints. We initiate the process with warm-up iterations on the original QA task to train the model. Following this, we continue the training while adding the constraint violation losses to the primary loss.

There are multiple methods of incorporating constraints into loss. We utilize a training algorithm inspired by~\cite{NEURIPS2019_cf708fc1}, designed for logical constraints, which we alter to apply to our probabilistic constraints. This method keeps the underlying architecture of the model the same, allowing us to transfer this model to other domains. It also assigns Lagrangian Multipliers $\lambda$  to each rule, which signifies its difficulty during training. While there are variations and heuristics for including constraint violation in the loss, such as \cite{li-etal-2019-logic}, we found the employed version a more principled way of implementing the optimization objective. We explain our definition of constraints in this method and our unique way of formulating them as a part of the loss, but the details of the rest of the optimization algorithm are not our contribution and, thus, are not discussed here. We refer the reader to see Appendix~\ref{sec:appendix_Pseudo_Code} for details of the training algorithm. As per the methodology outlined in~\cite{NEURIPS2019_cf708fc1}, we apply the dual formulation of the objective as follows, 

\begin{small}
\begin{equation}Loss = TaskLoss +\sum_{i=1} ^{m}\lambda_j*C_i, 
\label{eqn:loss}
\setlength{\abovedisplayskip}{1pt}
\setlength{\belowdisplayskip}{1pt}
\end{equation}
\end{small}

\noindent where ``TaskLoss''  denotes the primary task loss aiming to minimize the predicted probability error for the hypothesis. The new additional term is the constraint violation loss used in its dual form with Lagrangian multipliers, $\lambda_j$, where $j$ is the index of rule $j$ used in constraint violation $i$ ($C_i$). $m$ is the number of selected constraints. $\lambda_j$ is adjusted during training and ultimately indicates a rule's propensity to violation. Consequently, as training progresses, the loss function predominantly impacts the rules with the highest accumulated $\lambda_j$. 

\noindent \textbf{Inference} During inference, the model receives the context that includes textual rules and facts, while the formal rules and constraints that were employed during training, are not available to the model. We expect the model to learn to obey the rules that were utilized in the loss function during training. This ensures the model's generalizability and transferability across various domains.

\section{Dataset Creation}
\label{datasetcreation}

\noindent \textbf{Motivation} RuleTaker-pro is created to address some of the shortcomings of the RuleBERT dataset. RuleBERT~\cite{saeed-etal-2021-rulebert} is built using about 100 rules with fixed probabilities that are applied to many examples in the dataset. The probabilities of these rules are extracted from an external source and remain constant for all examples in the dataset. However, we want a dataset with example-specific rules to make the required reasoning more realistic. For example, the probability of two married people being cousins in the context of one culture is high, while it is close to zero in another or, in the medical domain, the prevalence or mortality of a disease varies depending on gender or location~\cite{zirra2023gender,menotti2023cardiovascular}. 

\noindent \textbf{Rule Generation} We developed RuleTaker-pro by modifying RuleTaker's crisp logical rules $(p_1,p_2,...,p_n) \to q$ (with $Pr$ equal to $1.0$) to include probabilities while the rest of the context remains unchanged (examples shown in the right side of Table~\ref{table:examples}). We leverage a Gaussian random generator to produce probabilities. The mean and variance of the Gaussian generator depend on the depth of reasoning to ensure a balanced dataset with a mean probability of 0.50 and an equal number of answers above and below $0.5$ probability. After assigning probabilities to the rules, we use Problog~\cite{10.5555/1625275.1625673}, a probabilistic logical inference tool that facilitates the encoding of probabilistic facts and rules, to compute the probability of the hypothesis. The resulting rules are similar to RuleBERT rules $(p_1,p_2,...,p_n) \to q, Pr$. See Appendix~\ref{sec:appendix_RuleTakerproGeneration} for details of data creation and distribution, which demonstrates its robustness.

\noindent \textbf{Adverbs of Uncertainty} In the context, we include the probability of the rule as an adverb of uncertainty like \emph{Usually}, \emph{Normally}, and \emph{Seldom} with associated probabilities of $0.90$, $0.80$, and $0.15$, respectively. A key difference between RuleTaker-pro and RuleBERT is including instance-specific rules. For example, the rule ``If A is a cousin of B, then A is a spouse of B.'' from RuleBERT will always have the probability of $0.15$ in all the examples. However, in Ruletaker-pro, the same rule may hold different probabilities depending on the adverb assigned to it in different instances. A rule such as ``Usually, if someone is big, then they are green.'' carries a probability of $0.90$ in one context, while  ``Seldom, if someone is big then they are green.'' carries a probability of $0.15$ in some other context. Given this difference, the model has to extract the rules from each context and can not use the information learned about the rules from the training data. See Appendix~\ref{sec:appendix_Adverbs} for more details.

\noindent \textbf{Metadata} Metadata about the inference of all facts and their depths are in the dataset and will be used to create constraints which would be used to train our model in PCT during training but \textbf{are not directly used during training or inference}. Notably, ambiguity and cycles have already been removed from the RuleTaker dataset for the logical rules and are not an issue in our dataset, as confirmed by our ProbLog solver. In addition, 20\% of examples in RuleTaker had a \textit{Complex} inference architecture, a ratio which we will keep as well.

\section{Experiments}
\label{sec:experiments}
In this section, we address four questions using our synthesized RuleTaker-pro and the RuleBERT datasets: \textbf{Q1}. How do textual rules affect probabilistic reasoning (\ref{sec:exp1})? We will also discuss the baseline results this section.
\textbf{Q2}. To what extent does the baseline language model improve with PCT concerning probabilistic reasoning and intermediate inferred facts (\ref{sec:exp2})? We also include the ablation study to investigate the impact of various losses and datasets on our approach using multiple metrics. \textbf{Q3}. Can we transfer the probabilistic reasoning capabilities of the language model when pre-trained with PCT(\ref{sec:transfer})? \textbf{Q4}. How do LLMs compare to fine-tuned BERT-based models (\ref{sec:LLMresult})? 

\noindent\textbf{Evaluation Metrics.}  We use several performance measures following~\cite{saeed-etal-2021-rulebert}. Binary Accuracy (BA) deems predictions correct if ground truth and predicted probability both fall under or over $0.5$. The CA25, CA10, and CA1 require the predicted probability to be in a window of $\pm0.25$, $\pm0.10$, and $\pm0.01$ of the ground truth, respectively. \cite{saeed-etal-2021-rulebert} applies CA10 and CA1 metrics to dataset splits with isolated rules, while BA is used for all reasoning depths for datasets involving all the rules. For comparison, we use BA for RuleBERT, but we thoroughly evaluate RuleTaker-pro using all relevant criteria. We use an extra metric, CS, to measure soft \textbf{Constraint Satisfaction} that deems the constraint (defined in Equation~\ref{eqn:constraint1}) satisfied if the following inequality holds:

\begin{small}
\begin{equation}| P(q) - P(p_1)*P(p_2)*...*P(p_n)*Pr | < Threshold.
\label{metric:pd1}
\setlength{\abovedisplayskip}{1pt}
\setlength{\belowdisplayskip}{4pt}
\end{equation}
\end{small}

\noindent This means that the difference between the predicted and calculated probability of an inferred fact, based on premise facts, must be less than a threshold. This threshold is $0.01$ for CS1, $0.10$ for CS10, and $0.25$ for CS25.

\subsection{Q1: Effect of Rules in Textual Format}
\label{sec:exp1}

\noindent \textbf{RuleBERT} Firstly, we investigate whether RoBERTa utilizes the text of the rule in the RuleBERT dataset by keeping and removing them from the context in two different experiments. For example, if we remove the textual rules in Table~\ref{table:examples}, the input will only include Facts 1 and 2. We report the results of these two settings in Table~\ref{tab:RulerBert_mybase}, where columns indicate the maximum depth of reasoning in training (M1-M5), and rows correspond to the reasoning depth of testing (D1-D5). We omit M0 as depth 0 does not use any rules, making it irrelevant to our investigation of PCT. We observe that the accuracy improves across most models and depths when the rules' text is excluded, suggesting that RoBERTa is \textbf{not using} it, and including it may even add unnecessary complexity. Thus, we conjecture that in RuleBERT dataset, RoBERTa can implicitly learn the probabilities of these rules from the facts and hypothesis in training data alone without using the textual rules explicitly.

\begin{table}[h]
\begin{center} \small

\begin{tabular}{|p{0.4cm}|p{0.8cm}|p{0.8cm}|p{0.8cm}|p{0.8cm}|p{0.8cm}|}
\hline
&\multicolumn{5}{c|}{Roberta With Text of the Rules}\\
\hline
 & M1 & M2 & M3 & M4  & M5 \\ 
\hline
D1  & \HeatCellRuleBERTMain{76.9}& \HeatCellRuleBERTMain{79.8}  &  \HeatCellRuleBERTMain{79.9} & \HeatCellRuleBERTMain{70.7}  &  \HeatCellRuleBERTMain{64.9} \\ 
D2  & \HeatCellRuleBERTMain{77.5}& \HeatCellRuleBERTMain{77.8}  & \HeatCellRuleBERTMain{76.6} & \HeatCellRuleBERTMain{70.4}  &  \HeatCellRuleBERTMain{65.4} \\ 
D3  & \HeatCellRuleBERTMain{78.4}& \HeatCellRuleBERTMain{76.9}  & \HeatCellRuleBERTMain{76.2}  & \HeatCellRuleBERTMain{78.8} & \HeatCellRuleBERTMain{71.6} \\
D4  & \HeatCellRuleBERTMain{76.2}& \HeatCellRuleBERTMain{73.4} & \HeatCellRuleBERTMain{72.4}  & \HeatCellRuleBERTMain{78.2} &  \HeatCellRuleBERTMain{73.8}  \\ 
D5  & \HeatCellRuleBERTMain{77.1}& \HeatCellRuleBERTMain{73.0}  & \HeatCellRuleBERTMain{69.6}  & \HeatCellRuleBERTMain{77.5}  &  \HeatCellRuleBERTMain{78.1} \\ 
\hline
&\multicolumn{5}{c|}{Roberta Without Text of the Rules}\\
\hline
D1  & \HeatCellRuleBERTMain{76.8}& \HeatCellRuleBERTMain{82.0} & \HeatCellRuleBERTMain{82.2} & \HeatCellRuleBERTMain{83.6}  & \HeatCellRuleBERTMain{82.1} \\ 
D2  & \HeatCellRuleBERTMain{75.4} & \HeatCellRuleBERTMain{78.8}   & \HeatCellRuleBERTMain{78.2} & \HeatCellRuleBERTMain{80.0} & \HeatCellRuleBERTMain{78.5} \\ 
D3  & \HeatCellRuleBERTMain{77.9} & \HeatCellRuleBERTMain{80.6}  & \HeatCellRuleBERTMain{80.6}  & \HeatCellRuleBERTMain{82.8}  & \HeatCellRuleBERTMain{80.6} \\
D4  & \HeatCellRuleBERTMain{75.0} & \HeatCellRuleBERTMain{76.2}  & \HeatCellRuleBERTMain{77.2}  & \HeatCellRuleBERTMain{79.6}  & \HeatCellRuleBERTMain{77.0}\\ 
D5  & \HeatCellRuleBERTMain{78.4} & \HeatCellRuleBERTMain{75.2}  & \HeatCellRuleBERTMain{78.7} & \HeatCellRuleBERTMain{79.6}   & \HeatCellRuleBERTMain{76.7}\\ 
\hline
&\multicolumn{5}{c|}{Roberta + PCT}\\
\hline
D1  & \HeatCellRuleBERTMain{79.1} & \HeatCellRuleBERTMain{81.7} & \HeatCellRuleBERTMain{82.4}  & \HeatCellRuleBERTMain{84.1}  & \HeatCellRuleBERTMain{81.1} \\ 
D2  & \HeatCellRuleBERTMain{78.5}& \HeatCellRuleBERTMain{79.7}  & \HeatCellRuleBERTMain{77.3}  & \HeatCellRuleBERTMain{80.9} & \HeatCellRuleBERTMain{77.7} \\
D3  & \HeatCellRuleBERTMain{79.8}& \HeatCellRuleBERTMain{83.4} & \HeatCellRuleBERTMain{81.9}  & \HeatCellRuleBERTMain{86.2}  & \HeatCellRuleBERTMain{82.2}\\
D4  & \HeatCellRuleBERTMain{77.4}& \HeatCellRuleBERTMain{81.4} & \HeatCellRuleBERTMain{80.2}  & \HeatCellRuleBERTMain{85.1}  & \HeatCellRuleBERTMain{81.3}\\ 
D5  & \HeatCellRuleBERTMain{80.1}& \HeatCellRuleBERTMain{84.3}  & \HeatCellRuleBERTMain{84.3} &  \HeatCellRuleBERTMain{86.1}  &  \HeatCellRuleBERTMain{83.6}\\
\hline
\end{tabular}

\end{center}

\caption{BA results of RoBERTa fine-tuned on RuleBERT. Columns indicate the maximum depth of reasoning in training (M1-M5), and rows correspond to the reasoning depth of testing (D1-D5). The results are shown for three different training settings: Roberta With Text of the Rules, Roberta Without Text of the Rules and Roberta + PCT.}
\label{tab:RulerBert_mybase}
\end{table}

Our baseline differs from~\cite{saeed-etal-2021-rulebert}. This discrepancy arises from our approach of freezing 22 transformer layers for faster training and more fine-tuned hyper-parameters, which yield superior accuracy at higher depths (We use the same loss function, Weighted binary cross-entropy). Moreover, we also train our models with~\cite{saeed-etal-2021-rulebert} original setting, and again, the text of the rules did not yield any positive impact on the performance (see Appendix~\ref{sec:appendix_originalrulebert} for details).

\noindent \textbf{RuleTaker-pro} We compare baseline results for the RuleTaker-pro dataset using Cross-Entropy (CE) and MSE loss functions for CA1 and CA10 metrics in Table~\ref{tab:ruletakerbase_losses}. Here, the Weighted binary cross-entropy was abandoned due to underperformance on RuleTaker-pro. The models are trained with maximum depths 1, 2, 3, and 5 (max), as these are the depths provided in the original RuleTaker training data. However, the testing is done on all depths 1 to 5, and their average accuracy is shown according to CA1 and CA10. $CS$ also averages over all depths. Though MSE excels in CA10, it underperforms in CA1 and CS1, especially when trained at higher depths. Our investigation into multiple cases indicates that MSE's low CS1 results from the minor MSE  approximation errors at lower depths, magnified at higher depths when multiplied along the chain of probabilities. 

\begin{table}[h]
\begin{center}
\small
\begin{tabular}{|p{0.7cm}|p{0.9cm}|p{0.6cm}|p{0.6cm}|p{0.6cm}|p{0.8cm}|}

 \hline
Loss & Metric & M1 & M2 & M3  & Mmax   \\ 
\hline
\multirow{2}{*}{CE} &CA1 & \HeatCellLosses{38.2} &  \HeatCellLosses{38.3} & \HeatCellLosses{20.4} & \HeatCellLosses{33.8}  \\ 
\cline{2-6}
 &CS1  & \HeatCellLosses{47.8}  & \HeatCellLosses{35.7}  & \HeatCellLosses{16.2} & \HeatCellLosses{20.7}   \\ 

\hline
\multirow{2}{*}{MSE}&CA1 & \HeatCellLosses{30.3} &  \HeatCellLosses{32.2} &\HeatCellLosses{26.1} & \HeatCellLosses{26.0}\\
\cline{2-6}
& CS1  & \HeatCellLosses{25.2} &  \HeatCellLosses{14.8}& \HeatCellLosses{14.4} & \HeatCellLosses{12.9} \\
\hline
\multirow{2}{*}{CE}& CA10 & \HeatCellLosses{46.4} &  \HeatCellLosses{49.6} & \HeatCellLosses{49.9} & \HeatCellLosses{53.2}\\
\cline{2-6}
&CS10  & \HeatCellLosses{52.2}  & \HeatCellLosses{44.9} & \HeatCellLosses{35.6} & \HeatCellLosses{38.2} \\
\hline
\multirow{2}{*}{MSE}& CA10 & \HeatCellLosses{58.1} &  \HeatCellLosses{62.7} & \HeatCellLosses{66.6} & \HeatCellLosses{74.8} \\
\cline{2-6}
&CS10  & \HeatCellLosses{45.1} &  \HeatCellLosses{34.4} & \HeatCellLosses{32.8} & \HeatCellLosses{33.3}  \\ 
\hline

\end{tabular}
\end{center}

\caption{Baseline RoBERTa's results for the RuleTaker-pro dataset using Cross-Entropy (CE) and MSE loss functions for CA1 and CA10 metrics. The models are trained with maximum depths 1, 2, 3, and 5 (max). The average accuracy of questions of all depths is shown according to CA1 and CA10. $CS$ shows constraint satisfaction as an average over all depths.}
\label{tab:ruletakerbase_losses}
\end{table}

Given our goal of achieving exact inference probabilities following the path of reasoning, CA1 is a more relevant measure for PCT evaluation. Also, given CE's higher CA1 and CS1 performance, we will focus mainly on CE and CA1's results which are detailed in Table~\ref{tab:ruletakerbase}. Detailed results for all losses and depths for metrics CA1, CA10, BA, MSE, and L1  are available in Appendix~\ref{sec:appendixRuleTaker_detailed}.

\begin{table}[h]
\small
\begin{center}
\begin{tabular}{|c|c|c|c|c|}
\hline
 &\multicolumn{4}{c|}{RoBERTa}\\
 \hline
D\slash M & M1 & M2 & M3  & Mmax   \\ 
\hline
Total & \HeatCell{38.2} &  \HeatCell{38.3} & \HeatCell{20.4} & \HeatCell{33.8}  \\ 
\hline
D1  & \HeatCell{56.0} &  \HeatCell{52.7} & \HeatCell{29.6} & \HeatCell{43.7}  \\ 
D2  & \HeatCell{36.4}  & \HeatCell{38.2} & \HeatCell{20.3} & \HeatCell{32.8}   \\ 
D3  & \HeatCell{29.3}  & \HeatCell{31.3}  & \HeatCell{14.9} & \HeatCell{28.3}   \\
D4  & \HeatCell{27.4}  & \HeatCell{28.5} & \HeatCell{14.0} & \HeatCell{27.1}   \\ 
D5  & \HeatCell{24.9}  & \HeatCell{26.7} & \HeatCell{14.7} & \HeatCell{28.2}   \\ 
\hline
CS1  & \HeatCell{47.8}  & \HeatCell{35.7}  & \HeatCell{16.2} & \HeatCell{20.7}   \\ 
\hline
&\multicolumn{4}{c|}{RoBERTa + PCT}\\
\hline
Total & \HeatCell{38.0} & \HeatCell{39.5} &  \HeatCell{41.1} &  \HeatCell{37.6}\\ 
 \hline
D1 & \HeatCell{53.3} &  \HeatCell{50.8} &  \HeatCell{50.5} &  \HeatCell{46.9}\\ 
D2 & \HeatCell{37.4} & \HeatCell{40.4} &  \HeatCell{42.2} &  \HeatCell{37.0}\\ 
D3 & \HeatCell{26.4} & \HeatCell{32.9} &  \HeatCell{36.0} &  \HeatCell{32.4}\\ 
D4 & \HeatCell{26.5} & \HeatCell{31.9} &  \HeatCell{33.9} &  \HeatCell{31.8}\\ 
D5 & \HeatCell{23.3} & \HeatCell{30.4} &  \HeatCell{33.4} &  \HeatCell{31.4}\\ 
\hline
CS1  & \HeatCell{44.9}  & \HeatCell{42.6} & \HeatCell{34.5} & \HeatCell{35.2} \\ 
\hline

\end{tabular}
\end{center}

\caption{Results of RoBERTa fine-tuned on RuleTaker-pro with CE loss, according to CA1 metric. Columns indicate the maximum depth of reasoning in training (M1-Mmax), and rows correspond to the reasoning depth of testing (D1-D5). The bottom section shows the improved results after the incorporation of PCT.}
\label{tab:ruletakerbase}
\end{table}

Unlike RuleBERT, RuleTaker-pro uses example-specific rules, \textbf{requiring the text of the rules} to determine the answer. Without rules, the predictions of our model are not better than random guesses. In RuleTaker-pro, we initially generated probabilistic rules by including the probability in the text, such as "With the probability of 15\%, if someone is green, then they are sad". However, we also considered using adverbs of uncertainty \cite{farkas-etal-2010-conll} instead of numbers, changing the rule to "Seldom, if someone is green, then they are sad". Adverbs of uncertainty improved the models in Dev BA by 0.5\%-2\%, thus we followed this approach in RuleTaker-pro creation (see Appendix~\ref{sec:appendix_Adverbs}).

\subsection{Q2: Effectiveness of PCT}
\label{sec:exp2}

\noindent \textbf{RuleBERT} Table~\ref{tab:RulerBert_mybase} displays the impact of PCT on improving RuleBERT's accuracy over the baseline results of RoBERTa, especially at deeper depths. These results are for the baseline without the text of the rules (results with the rule's text yield similar outcomes; see Appendix~\ref{sec:appendix_RuleBERT_with_text}). Using PCT, the CS25 accuracy of intermediate inferred facts increases from an average of 50\% to over 90\%. Increasing the constraint satisfaction of intermediate inferred facts works synergistically with the accuracy of the model by compelling the model to reason, thus, enhancing it, especially at deeper depths. Appendix~\ref{sec:appendixIIF} includes more details about inferred intermediate facts.

\noindent \textbf{RuleTaker-pro} By deploying PCT in RuleTaker-pro, we observe a similar trend to RuleBERT. As illustrated in Table~\ref{tab:ruletakerbase}, by incorporating exact probabilities into the constraints, PCT improves the accuracy of CA1 in most models. Another place where PCT shows improved generalization is when it is used to train the models at lower depths, i.e., 2 and 3, and tested at higher depths. This shows that the reasoning learned with PCT is transferred to higher depths. However, at depth 1, due to the limited number of applicable constraints, the change in accuracy is minor. Similar to RuleBERT, we observe a sharp increase of about 50\% in the CS in all the models trained with PCT.

\noindent\textbf{Error Analysis.} Our findings indicate that improvements in constraint consistency are not always proportionate to improvements in accuracy. This discrepancy is prevalent in nearly all tasks involving constraints, as evidenced by related studies~\cite{ribeiro-etal-2019-red}. Notably, to maintain the consistency of outputs, the model might yield incorrect results. Incorporating PCT encouraged the model to output lower probabilities than the baseline model, thus reducing the magnitude of the constraint loss. For instance, in the model trained at depth 3 with PCT, the average output probabilities for all the test dataset questions declined from a baseline of 52\% to 45\%. When the model is trained with depth 1 with PCT, the constraint satisfaction decreases, likely due to its reduced ability to accurately process questions with a higher reasoning depth. In short, while the best results are achieved when both CS and CA increase, a high CS does not invariably guarantee a corresponding increase in CA. See Appendix~\ref{sec:appendixerroranalysis} for detailed examples.

\subsection{Q3: Transferability Analysis}
\label{sec:transfer}

Experiments in Section~\ref{sec:exp2} highlighted the effectiveness of PCT in transferring \textit{reasoning} from a model trained at lower depths to answer questions at higher depths. Here, we evaluate the transferability of PCT from different perspectives. 

\noindent\textbf{Transferring Reasoning From Simple to Complex Examples.} As highlighted in Section~\ref{datasetcreation}, 20\% of the inference questions in RuleTaker-pro have the \textit{complex} architecture. Table~\ref{tab:SVSC_CE_MSE} presents our models' performance on \textit{simple} and \textit{complex} questions separately, with the models predictably faring better on the former. Employing CE+PCT increases accuracy for both question types, making the difference between them negligible. This suggests that the models can do probabilistic reasoning even in \textit{complex} instances. However, for MSE and MSE+PCT models, the difference between the performance over different question types remains substantial. Using PCT along with cross-entropy loss in the CE+PCT model was more effective in learning probabilistic reasoning because PCT directs the model to output the exact probability values that do not violate the rules. However, the MSE model does not see the same benefit due to cascading errors in the approximated probabilities of the inferred facts, as discussed in Section~\ref{sec:exp2}. In the case of MSE, adding PCT still improves accuracy.

\begin{table}[h]
\begin{center} 
\small
\begin{tabular}{|p{0.2cm}||p{0.6cm}|p{0.6cm}|p{0.8cm}||p{0.6cm}|p{0.6cm}|p{0.8cm}||}
 \hline
&\multicolumn{3}{c||}{CE}&\multicolumn{3}{c||}{CE+PCT}\\
 \hline
 &  M2 & M3  & Mmax  & M2 & M3  & Mmax  \\ 
\hline
S & \HeatCellGPT{39} &  \HeatCellGPT{20} & \HeatCellGPT{34} & \HeatCellGPT{41} & \HeatCellGPT{40} &  \HeatCellGPT{37}   \\ 
\hline
C & \HeatCellGPT{34} &  \HeatCellGPT{18} & \HeatCellGPT{32} & \HeatCellGPT{36} & \HeatCellGPT{38} &  \HeatCellGPT{36}   \\ 
\hline
\hline
&\multicolumn{3}{c||}{MSE}&\multicolumn{3}{c||}{MSE+PCT}\\
 \hline
 &  M2 & M3  & Mmax  & M2 & M3  & Mmax  \\ 
\hline
S & \HeatCellGPT{33} & \HeatCellGPT{27} & \HeatCellGPT{27} & \HeatCellGPT{36} & \HeatCellGPT{37} &  \HeatCellGPT{35}   \\ 
\hline
C & \HeatCellGPT{24} & \HeatCellGPT{19} & \HeatCellGPT{20}  & \HeatCellGPT{27} & \HeatCellGPT{28} &  \HeatCellGPT{30}   \\ 
\hline
\end{tabular}

\end{center}
\caption{RuleTaker-pro results on Simple (S) and Complex (C) examples trained with Cross Entropy and MSE, before after addition of PCT. }
\label{tab:SVSC_CE_MSE}
\end{table}

\noindent\textbf{Domain Transfer.} We evaluated the transferability of the probabilistic reasoning and constraint satisfaction capabilities to another domain by training our model on RuleTaker-pro with CE+PCT and fine-tuning it on RuleBERT. This transfer direction is selected due to the superior constraint satisfaction of the model that was trained on the RuleTaker-pro dataset. We compare the RuleBERT baseline with two transfer learning approaches: 1) Pre-training RoBERTa on the RuleTaker-pro dataset with a simple CE loss (Augmented Data) to ensure the improvements are not the result of increased data alone, 2) Pre-training RoBERTa on RuleTaker-pro dataset with CE+PCT loss (Transfer Learning of PCT), aiming to understand the specific impact of pre-training with PCT. The findings, detailed in Table~\ref{tab:DomainTransfer}, show that only lower-depth results improved by data augmentation, while higher depths and overall accuracy improved by transferring from CE+PCT. Transferring from the CE+PCT also increased CS measures for depths 2, 3, and 5 by about +4, +14, and +7, respectively. In contrast, using Augmented Data did not result in any changes in CS measures.

\begin{table}[!h]
\begin{center} 
\small
\begin{tabular}{|p{0.4cm}|p{0.8cm}|p{0.8cm}|p{0.8cm}|}
 \hline
 &\multicolumn{3}{c|}{Baseline RoBERTa}\\ 
 \hline
 & M2 & M3 & M5 \\ 
 \hline

 D2 & \HeatCellRuleBERT{77.8} & \HeatCellRuleBERT{76.6} & \HeatCellRuleBERT{65.4}\\ 
 D3 & \HeatCellRuleBERT{76.9} & \HeatCellRuleBERT{76.2} & \HeatCellRuleBERT{71.6}\\
 D4 & \HeatCellRuleBERT{73.4} & \HeatCellRuleBERT{72.4} & \HeatCellRuleBERT{73.8}\\ 
 D5 & \HeatCellRuleBERT{73.0} & \HeatCellRuleBERT{69.6} & \HeatCellRuleBERT{78.1}\\ 
 \hline
 &\multicolumn{3}{c|}{Augmented Data}\\ 

 \hline
 D2 & \HeatCellRuleBERT{76.8} & \HeatCellRuleBERT{80.6} & \HeatCellRuleBERT{83.4}\\ 
 D3 & \HeatCellRuleBERT{75.9} & \HeatCellRuleBERT{83.2} & \HeatCellRuleBERT{81.6}\\
 D4 & \HeatCellRuleBERT{70.4} & \HeatCellRuleBERT{76.4} & \HeatCellRuleBERT{74.8}\\ 
 D5 & \HeatCellRuleBERT{68.0} & \HeatCellRuleBERT{72.6} & \HeatCellRuleBERT{67.1}\\ 
 \hline
 &\multicolumn{3}{c|}{Transfer Learning of PCT}\\ 

 \hline
 D2 & \HeatCellRuleBERT{84.8} & \HeatCellRuleBERT{84.6} & \HeatCellRuleBERT{72.4}\\ 
 D3 & \HeatCellRuleBERT{84.9} & \HeatCellRuleBERT{82.2} & \HeatCellRuleBERT{72.6}\\
 D4 & \HeatCellRuleBERT{84.4} & \HeatCellRuleBERT{77.4} & \HeatCellRuleBERT{73.8}\\ 
 D5 & \HeatCellRuleBERT{86.0} & \HeatCellRuleBERT{66.6} & \HeatCellRuleBERT{81.1}\\ 
 \hline
\end{tabular}

\end{center}

\caption{Improvements in the binary accuracy (BA) and constraints satisfaction of RuleBERT models in Table~\ref{tab:RulerBert_mybase} after transfer learning from RuleTaker-pro. The results include the Baseline RoBERTa, the Augmented Data model that is trained on RuleBERT and then finetuned on RuleTaker-pro and Transfer Learning of PCT.}
\label{tab:DomainTransfer}
\end{table}

\subsection{Q4: LLM Results}
\label{sec:LLMresult}

To evaluate LLMs, we add instructions and examples (for few-shot settings) to their prompts. The LLM results for RuleTaker-pro for CA1 are shown in Table~\ref{tab:LLMresults}. We observe that even GPT3.5 with few-shot examples and GPT4 fall short of RoBERTa's accuracy. GPT4 with few-shot is not included in the table since adding few-shot examples to GPT4 or using COT did not improve but hurt our model. A similar outcome is reported on a different dataset~\cite{Shi2022StepGameAN} where incorporation of COT either marginally helped the model or hurt its accuracy at different depths of reasoning for multi-hop spatial reasoning~\cite{yang-etal-2023-coupling}. We believe that COT can potentially improve the LLM results, but it requires a significant time investment in prompt engineering and example selection. LLMs are undermined even more after we add PCT and improve RoBERTa's results. The gap in accuracy becomes even wider if we use the CA10 metric, where the CA10 accuracies remain almost the same as in CA1. This indicates that if the LLM cannot predict the exact probability, its prediction will not be even close to the correct answer. The results of LLMs on the RuleBERT dataset are as poor as a random baseline in all settings, even with the addition of COT for GPT4. See Appendix~\ref{sec:appendixLLMRuleTakerpro} for the details of prompt instructions, RuleTaker-pro CA10, and RuleBERT results. We further discuss the underperformance of LLMs on these and similar datasets in the same section of the appendix.

\begin{table}[h]
\centering
\small
\begin{tabular}{|c|c|c|c|c|}
\hline
 & RoBERTa & GPT3.5 & GPT3.5* & GPT4  \\
\hline
D1 & \HeatCellGPT{44} & \HeatCellGPT{28} & \HeatCellGPT{41} & \HeatCellGPT{41}  \\
\hline
D2 & \HeatCellGPT{33} & \HeatCellGPT{20} & \HeatCellGPT{26} & \HeatCellGPT{27}  \\
\hline
D3 & \HeatCellGPT{28} & \HeatCellGPT{23} & \HeatCellGPT{25} & \HeatCellGPT{26} \\
\hline
D4 & \HeatCellGPT{27} & \HeatCellGPT{18} & \HeatCellGPT{20} & \HeatCellGPT{17}\\
\hline
D5 & \HeatCellGPT{28} & \HeatCellGPT{18} & \HeatCellGPT{20} & \HeatCellGPT{21}\\
\hline
\end{tabular}

\caption{LLM results on RuleTaker-pro for CA1 and CA10 metrics. * indicates using few-shot examples. The chosen RoBERTa model is M5 trained with CE since it performs the best regarding CA1.}
\label{tab:LLMresults}
\end{table}

\section{Conclusion and Future Work}

Addressing the problem of reasoning over uncertain rules in textual format, we create a new dataset, RuleTaker-pro, extending the limited resources for studying this problem. We investigate how uncertain rules can be represented in the text and used by the learning models. We propose a novel approach that explicitly uses the rules of probabilistic reasoning as constraints in the loss. This approach improves the performance and reasoning of the backbone language models. Our experiments on LLMs have revealed that they struggle to perform probabilistic reasoning in zero-shot and few-shot scenarios, despite their impressive capabilities in solving other NLP tasks. Our future objective is to develop models that utilize the text of the rules more effectively and transfer their reasoning abilities to more realistic QA domains featuring uncertainty and more advanced structures of probabilistic reasoning. Also, it is worth exploring prompt engineering methods for Large Language Models to ease the use of uncertain text and inference on them.

\section*{Acknowledgements}
This project is supported by the National Science Foundation (NSF) CAREER award 2028626 and partially supported by the Office of Naval Research (ONR) grant N00014-20-1-2005 and grant N00014-23-1-2417. Any opinions, findings, conclusions, or recommendations expressed in this material are those of the authors and
do not necessarily reflect the views of the National Science Foundation nor the Office of Naval Research. We thank all reviewers for their thoughtful
comments and suggestions.

\section*{Limitations}

One limitation of our work is the fixed structure of the rules in our datasets, which limits the model's transferability to other domains with more open forms of explaining probabilistic rules. Another limitation is that we take a small step to formalize probabilistic reasoning over text. However, this does not mean the outcome language models are fully capable of language understanding and reasoning. Finally, running our models, based on  RoBERTa large while possible, is computationally expensive, limiting their usage with all our different settings. This is exacerbated when it comes to utilizing Large Language Models that, in their current state, are very expensive to use even in zero-shot and few-shot settings.

\bibliography{anthology,custom}
\bibliographystyle{acl_natbib}

\appendix

\section{Appendix}
\subsection{Adverbs in RuleTaker-pro}
\label{sec:appendix_Adverbs}

In creation of RuleTaker-pro, we utilize 8 different adverbs of frequency shown in the Table~\ref{tab:adverb}. Using adverbs of frequency improved the Dev binary accuracy consistently in all depths. The results are shown in Table~\ref{tab:devaccuracy}.

\begin{table*}[h]
\begin{center}
\begin{tabular}{|c|c|c|c|c|c|c|c|c|}
 \hline
Adverbs & always & usually & normally & often & sometimes & occasionally & seldom & never\\ 
\hline
Probability & 1.00 & 0.90 & 0.80 & 0.65 & 0.50 & 0.30 & 0.15 & 0.0\\ 

\hline

\end{tabular}
\end{center}

\caption{The adverb of uncertainty and their respective probabilities that we link to them.}
\label{tab:adverb}
\end{table*}

\begin{table}[h]
\begin{center} \small
\begin{tabular}{|c|c|c|c|c|c|c|c|c|c}
 \hline
 CE & M1 & M2 & M3 & Mmax \\ 
 \hline
 With adverbs & 96.24 & 94.97 & 93.12 & 89.71 \\  
 \hline
 With probabilities & 95.78 & 93.71 & 92.52  & 88.01 \\  
\hline

\end{tabular}
\end{center}

\caption{Dev BA for models M1 to Mmax trained with CE loss.}
\label{tab:devaccuracy}
\end{table}

\subsection{RuleTaker-pro Generation Algorithm}
\label{sec:appendix_RuleTakerproGeneration}

In order to make a balanced dataset with an equal number of labels, we generate a random probability for each rule based on a Gaussian random generator. Then the adverb with the closest probability to the generated probability is chosen. The rule probability generations are generated so that half of the answers are above and half are below 0.50.

The algorithm to change a logical context to a probabilistic one is shown in Algorithm~\ref{alg:datasetcreationalgorithm}. ``FIND\_ADVERB'' function gets a random probability from 0 to 100 as input and returns an adverb to it based on the closest probability of an adverb in Table~\ref{tab:adverb}. In the procedure ``ADD\_PROBABLITIES'', a logical context and question are given as input. Then, in line 6, it is randomly decided whether or not the final answer to this instance should be above or below 0.50 to ensure balance in the final results of the dataset. In the rest of the algorithm, until the pre-selected above or below 0.50 probability for the answer is achieved, random probabilities would be assigned to the rules in the context. The random function that assigns these probabilities is a Gaussian function with a mean of 40 and std of 60. The random probabilities are added with the value h, initially set to $depth*10$, and it increases or decreases slightly to help achieve the desired answer after reaching failure. h is created based on the depth of the dataset group to create a balanced average of answer probabilities. A real example of the created dataset is shown and analyzed in section~\ref{sec:appendixerroranalysis}.

\begin{algorithm}
\caption{Assigning Gaussian-based probabilities to logical rules to crate a probabilistic dataset while ensuring that the resulting dataset is balanced with heuristics.}
\label{alg:datasetcreationalgorithm}
\begin{algorithmic}[1]
\Function{find\_adverb}{$x$}
    \State Determine adverb and its associated probability based on the range of $x$ in Table~\ref{tab:adverb}
    \State \Return adverb
\EndFunction
\Statex
\Procedure{add\_probablities}{$c, q, d$} \Comment{$c$ is context, $q$ is question and $d$ is the depth of the dataset group (not the instance)}
\State $Above0.50 \gets \Call{Random}{False,True}$
\State $h \gets 10*depth$
\While {not Answer is $Above0.50$}
    
    \State $new\_c= c$
    \For {each rule in $context$}
        \State $p_i$ = \Call{RandomGauss}{40,60}+h
        \State $adverb= \Call{find\_adverb}{p_i}$
        \State add $adverb$ to $new\_c$
    \EndFor
    \State $Answer \gets \Call{problog}{new\_c, q}$

    \If{$Above0.50$}
        \State $h = h + 5$
    \Else
        \State $h = h - 5$
    \EndIf
    
\EndWhile
\EndProcedure
\end{algorithmic}
\end{algorithm}

Statistics about the splits, their unique context and questions, and their balanced average answer produced by our algorithm are shown in Table~\ref{tab:RuleTakerStatistics}.

\begin{table}[ht]
\begin{center} 

\begin{tabular}{|c|c|c|c|c|}
\hline
\textbf{Split} & \textbf{D} & \textbf{Rows} & \textbf{Queries} & \textbf{MA} \\
\hline
Train & 1 & 13549 & 807 & 0.49 \\
Train & 2 & 16145 & 810 & 0.48 \\
Train & 3 & 19960 & 812 & 0.48 \\
Train & 5 & 23805 & 812 & 0.50 \\
\hline
Dev & 1 & 1946 & 551 & 0.50 \\
Dev & 2 & 2290 & 586 & 0.48 \\
Dev & 3 & 2837 & 629 & 0.48 \\
Dev & 5 & 3412 & 694 & 0.50\\
\hline
Test & 1 & 3930 & 690 & 0.49 \\
Test & 2 & 4592 & 718 & 0.48 \\
Test & 3 & 5687 & 765 & 0.48 \\
Test & 5 & 6829 & 789 & 0.50\\
\hline
\end{tabular}
\end{center}
\caption{RuleTaker-pro Dataset Statistics. Split determines the split of the dataset which could be train, dev or test. D determines the depth of the question. Rows shows the total rows and Queries shows the number of unique queries. MA is the Mean of all Answers which should be around 0.50 for a balanced dataset.}
\label{tab:RuleTakerStatistics}
\end{table}

RuleTaker-pro depth distribution for all depths and the number of True and False labels are shown in Table~\ref{tab:RuleTakerdepths}.

\begin{table}[h]
\begin{center}
\begin{tabular}{|c||c|c|c|c|}
\hline
\hline
& M1 & M2 & M3 & Mmax \\
\hline
\hline
D0 T & 10626 & 9590 & 7441 & 2616 \\
\hline
D0 F & 10719 & 9485 & 7650 & 2720 \\
\hline
D1 T & 6422 & 4613 & 4438 & 3802 \\
\hline
D1 F & 6452 & 4465 & 4272 & 3692 \\
\hline
D2 T & 0 & 3441 & 2930 & 2442 \\
\hline
D2 F & 0 & 3469 & 2949 & 2520 \\
\hline
D3 T & 0 & 0 & 2597 & 2118 \\
\hline
D3 F & 0 & 0 & 2642 & 2026 \\
\hline
D4 T & 0 & 0 & 0 & 1852 \\
\hline
D4 F & 0 & 0 & 0 & 1858 \\
\hline
D5 T & 0 & 0 & 0 & 1761 \\
\hline
D5 F & 0 & 0 & 0 & 1734 \\
\hline
\end{tabular}
\end{center}
\caption{RuleTaker-pro depth distribution for all depths and the number of True and False labels. M* shows the distribution for the training set of max depth *. T and F stand for True and False labels which would indicate a probability of final answer being higher or lower than 0.50. }
\label{tab:RuleTakerdepths}
\end{table}

\subsection{ProbLog}
\label{sec:appendix_problog}

\begin{figure}[!h]

    \begin{subfigure}{0.21\textwidth}
    \includegraphics[width=\linewidth]{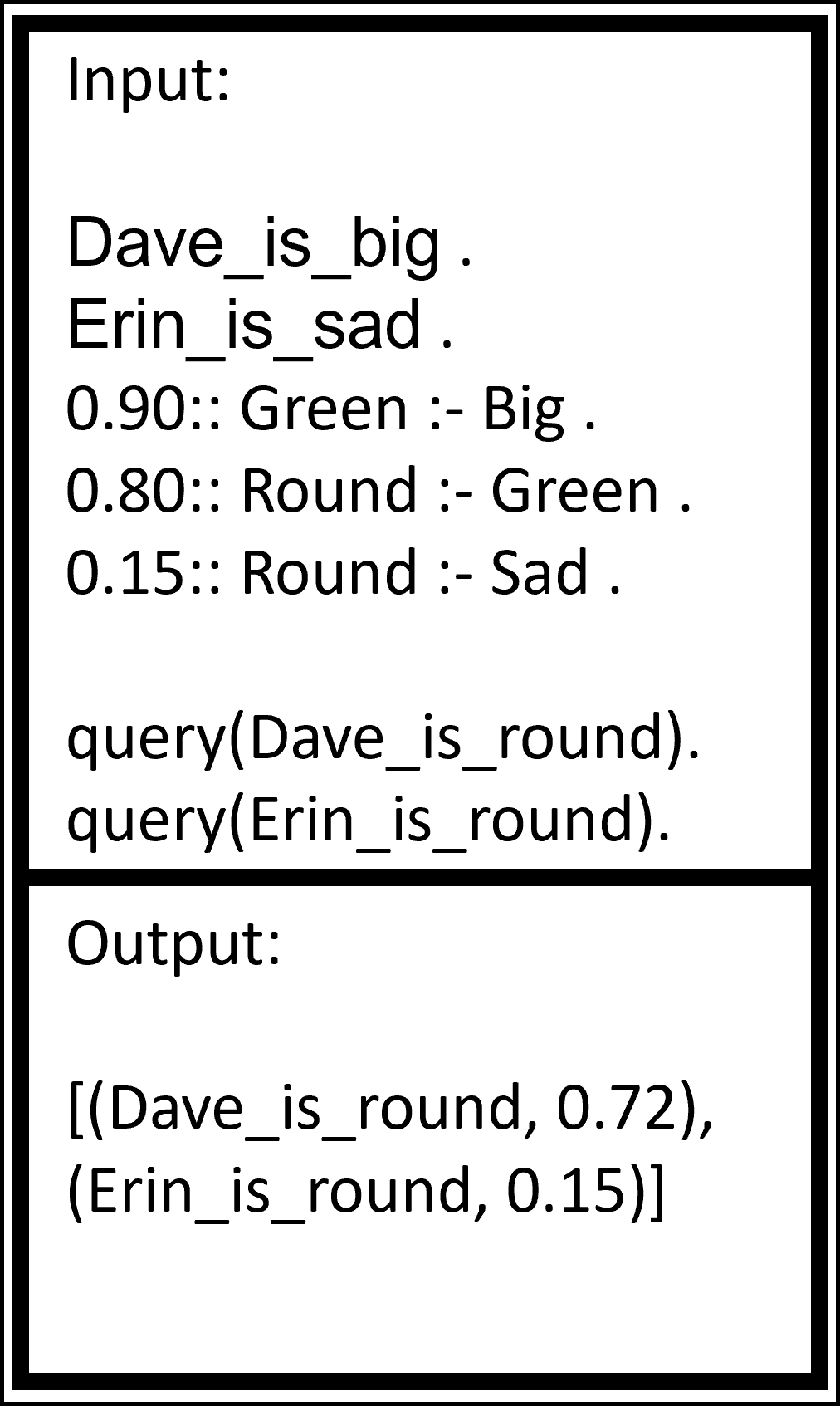}
    \caption{Encoding of the Table~\ref{table:examples}'s right column example in ProbLog pseudo code.} \label{fig:2a}
  \end{subfigure}%
  \hspace*{\fill}   
  \begin{subfigure}{0.20\textwidth}
    \includegraphics[width=\linewidth]{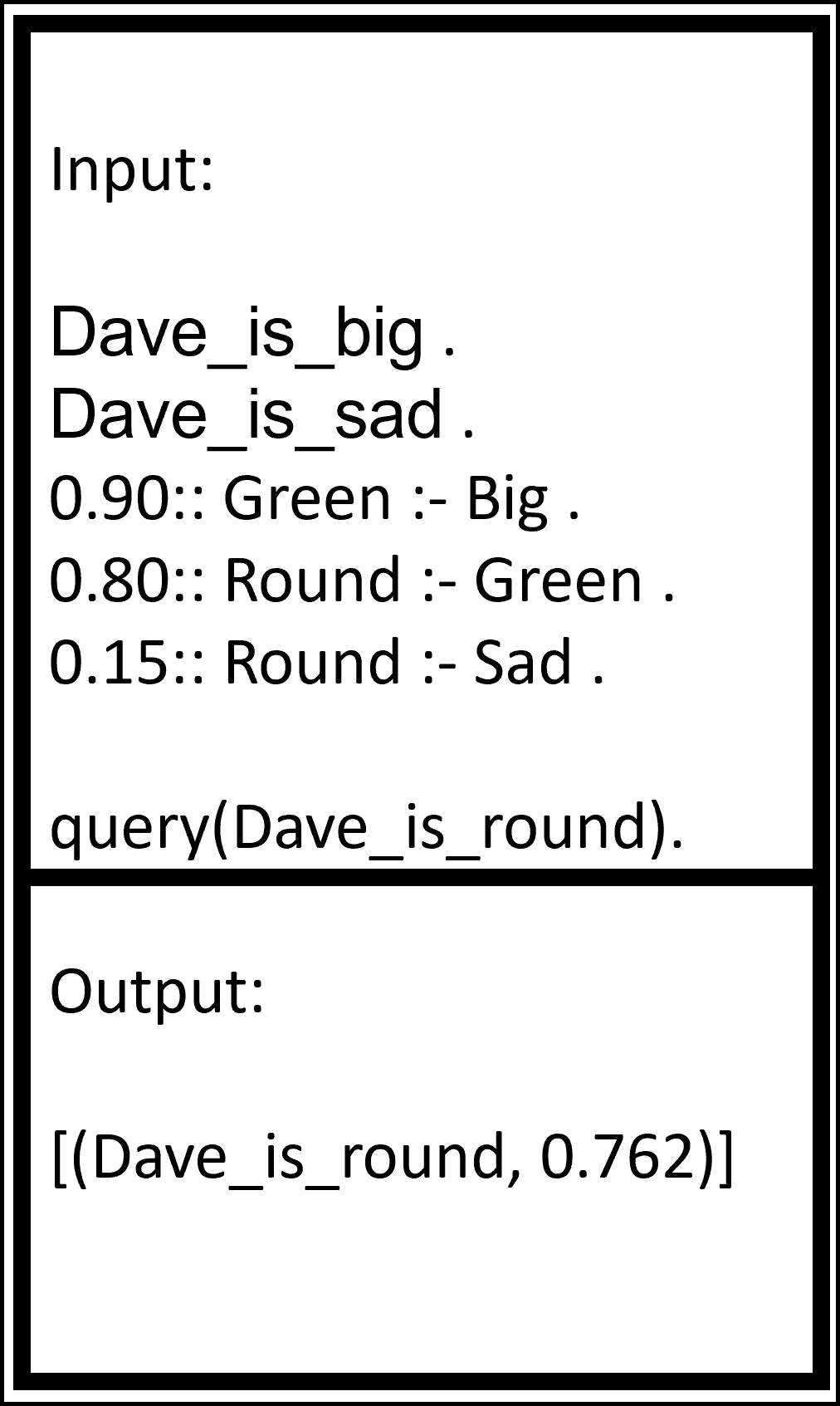}
    \caption{Encoding of the Table~\ref{table:examples}'s right column example in ProbLog pseudo code if the second fact is replaced with ``David is sad.''} \label{fig:2b}
  \end{subfigure}%

\end{figure}

ProbLog is a tool that allows us to encode probabilistic facts and rules. Then it will calculate any queries in the context of the defined facts and rules, which is exactly what we need for RuleTaker-pro. For example, Table~\ref{table:examples}'s right column would be shown in Problog pseudo code in the Figure~\ref{fig:2a}.

A more complicated example would occur when there is more than one way to reach an inferred intermediate fact. Imagine that the second fact in the example of Table~\ref{table:examples}'s right column is ``David is sad.''. In that case, the probability that ``David is round'' would be 0.762 as shown in Figure~\ref{fig:2b}.

\subsection{RuleBERT's Additional Results}
\subsubsection{RuleBERT's Original Setting}
\label{sec:appendix_originalrulebert}
The original RuleBERT baseline from~\cite{saeed-etal-2021-rulebert} is shown in Table~\ref{tab:RulerBert_Base}. We also train our models with their settings, both with and without including the text of the rules. These new results are shown in Table~\ref{tab:rulebertreview}. The text of the rules is still not useful for the models.

\begin{table}[h]
\begin{center} 

\begin{tabular}{|c|c|c|c|c|c|}
 \hline
 & M1& M2 & M3 & M4  & M5 \\ 
\hline
D1 & 86.0 & 88.4 & 88.7 & 88.9 & 88.9 \\ 
D2 & 65.5 & 73.0 & 75.1 & 75.0 & 72.0  \\ 
D3 & 58.1 & 63.6 & 68.4 & 69.0 & 65.6 \\
D4 & 46.8 & 54.7 & 62.6 & 66.6 & 62.7 \\ 
D5 & 35.6 & 49.6 & 70.3 & 78.5 & 74.4 \\ 
\hline

\end{tabular}
\end{center}

\caption{RuleBERT baseline results trained and tested on different depths \cite{saeed-etal-2021-rulebert}.}
\label{tab:RulerBert_Base}
\end{table}

\begin{table}
\begin{center} 

    \begin{tabular}{|c|c|c|c|c|c|}
        \hline
        &\multicolumn{5}{c|}{RoBERTa With Rules' Text}\\
        \hline
        D/M & M1 & M2 & M3 & M4 & M5 \\
        \hline
        D1  & 76 & 91 & 87 & 91 & 93 \\
        D2  & 76 & 87 & 79 & 83 & 83 \\
        D3  & 67 & 85 & 76 & 76 & 73 \\
        D4  & 66 & 82 & 69 & 63 & 51 \\
        D5  & 53 & 75 & 54 & 34 & 28 \\
        \hline
        &\multicolumn{5}{c|}{RoBERTa Without Rules' Text}\\
        \hline
        D1 & 88 & 90 & 88 & 92 & 89 \\
        D2 & 87 & 88 & 77 & 78 & 74 \\
        D3 & 84 & 85 & 73 & 72 & 67 \\
        D4 & 82 & 80 & 65 & 60 & 51 \\
        D5 & 80 & 68 & 44 & 29 & 21 \\
        \hline
    \end{tabular}
\end{center}
    \caption{M shows the maximum depth of the training data, and D shows the depth of the test data. Here RoBERTa is trained with the original~\cite{saeed-etal-2021-rulebert}'s training setting and parameters.}
    \label{tab:rulebertreview}
\end{table}

\subsubsection{PCT With Rules' Text}
\label{sec:appendix_RuleBERT_with_text}
Since the text of the rules decreases the accuracy of our models, we removed it in our original PCT result, but if we do include the text, PCT would still improve the accuracies as shown in Table~\ref{tab:RulerBert_PD_with_text}.

\begin{table}[h]
\begin{center} 
\begin{tabular}{|p{0.4cm}|p{0.6cm}|p{0.6cm}|p{0.6cm}|p{0.6cm}|p{0.6cm}|}
 \hline
&\multicolumn{5}{c|}{RoBERTa}\\
\hline
 & M1 & M2 & M3 & M4  & M5 \\ 
\hline
D1  & \HeatCellRuleBERT{76.9}& \HeatCellRuleBERT{79.8}  &  \HeatCellRuleBERT{79.9} & \HeatCellRuleBERT{70.7}  &  \HeatCellRuleBERT{64.9} \\ 
D2  & \HeatCellRuleBERT{77.5}& \HeatCellRuleBERT{77.8}  & \HeatCellRuleBERT{76.6} & \HeatCellRuleBERT{70.4}  &  \HeatCellRuleBERT{65.4} \\ 
D3  & \HeatCellRuleBERT{78.4}& \HeatCellRuleBERT{76.9}  & \HeatCellRuleBERT{76.2}  & \HeatCellRuleBERT{78.8} & \HeatCellRuleBERT{71.6} \\
D4  & \HeatCellRuleBERT{76.2}& \HeatCellRuleBERT{73.4} & \HeatCellRuleBERT{72.4}  & \HeatCellRuleBERT{78.2} &  \HeatCellRuleBERT{73.8}  \\ 
D5  & \HeatCellRuleBERT{77.1}& \HeatCellRuleBERT{73.0}  & \HeatCellRuleBERT{69.6}  & \HeatCellRuleBERT{77.5}  &  \HeatCellRuleBERT{78.1} \\ 
\hline
&\multicolumn{5}{c|}{RoBERTa + PCT}\\
\hline
D1 & \HeatCellRuleBERT{78.3} & \HeatCellRuleBERT{83.1} &  \HeatCellRuleBERT{77.5} & \HeatCellRuleBERT{77.9} &  \HeatCellRuleBERT{67.7} \\
D2 & \HeatCellRuleBERT{78.9} & \HeatCellRuleBERT{79.7} & \HeatCellRuleBERT{76.6} & \HeatCellRuleBERT{78.0} &  \HeatCellRuleBERT{68.9} \\
D3 & \HeatCellRuleBERT{79.1} & \HeatCellRuleBERT{80.8}  & \HeatCellRuleBERT{81.3} & \HeatCellRuleBERT{81.3} & \HeatCellRuleBERT{78.9}\\
D4 & \HeatCellRuleBERT{77.7} & \HeatCellRuleBERT{77.0} & \HeatCellRuleBERT{79.0} &  \HeatCellRuleBERT{80.8} &   \HeatCellRuleBERT{82.6} \\
D5 & \HeatCellRuleBERT{77.8} & \HeatCellRuleBERT{74.1} & \HeatCellRuleBERT{84.1} & \HeatCellRuleBERT{82.7} &  \HeatCellRuleBERT{88.6}\\
\hline
\end{tabular}

\end{center}
\caption{BA results of RoBERTa fine-tuned on RuleBERT when the rule's text is included with and without PCT. Columns indicate the maximum depth of reasoning in training (M1-M5), and rows correspond to the reasoning depth of testing (D1-D5). }
\label{tab:RulerBert_PD_with_text}
\end{table}

\subsection{CA25 Accuracy of Intermediate Inferred Facts}
\label{sec:appendixIIF}
\begin{figure}[h]
    \includegraphics[width=0.45\textwidth]{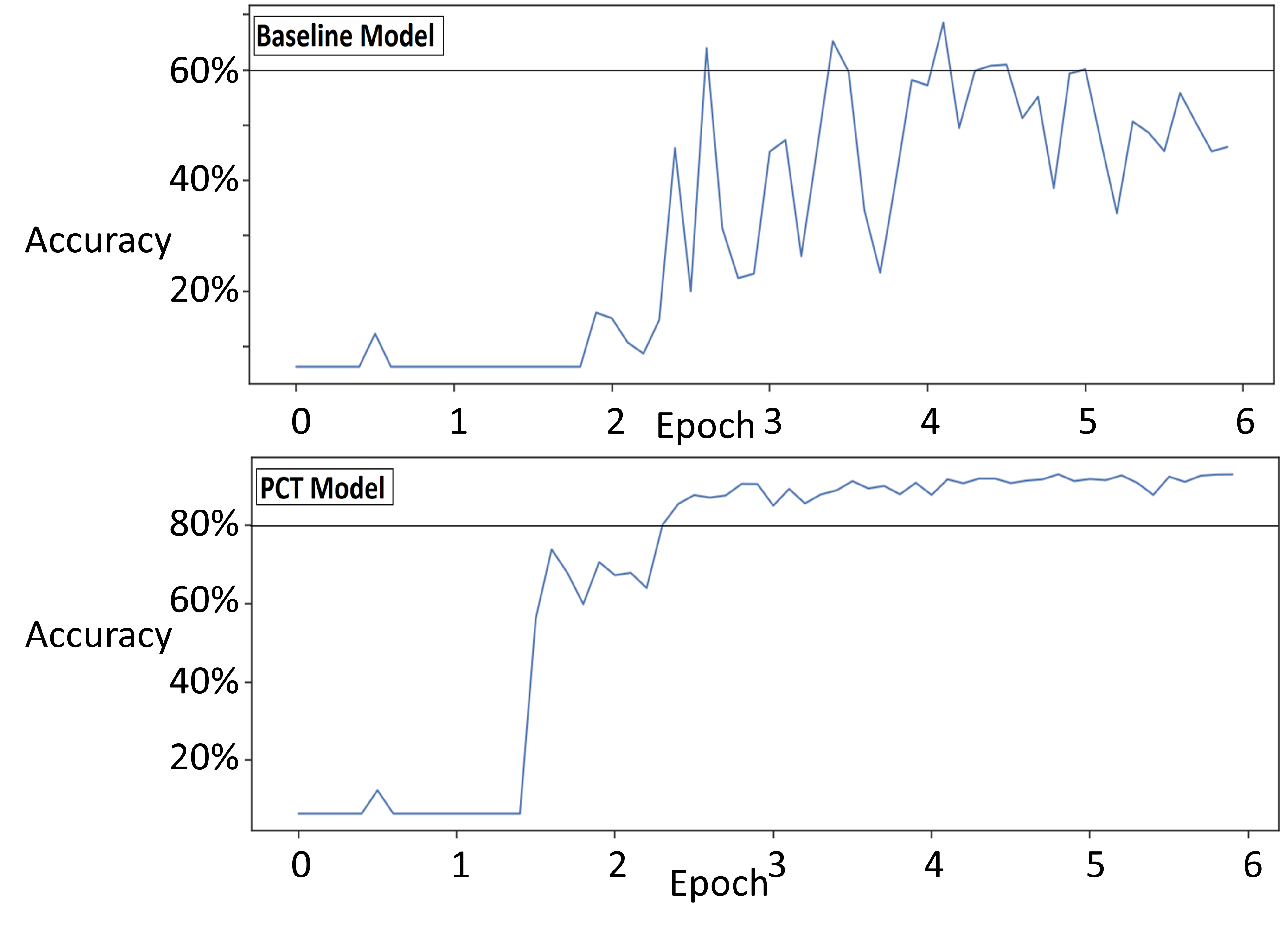}
    \caption{The CS25 of intermediate inferred facts over 6 Epochs of training for M5.}
    \label{fig:in_between1_AC}
\end{figure}

CA25 Intermediate Inferred Facts for M5 is depicted in Figure~\ref{fig:in_between1_AC}. The model is trained for 6 epochs to show the accuracy over time. PCT accuracy remains consistently over 0.90 while the baseline models accuracy fluctuates and remains below 0.60.

\subsection{PCT Algorithm Pseudo-Code}
\label{sec:appendix_Pseudo_Code}

The PCT algorithm pseudo-code is shown in Algorithm~\ref{alg:PCTalgorithm}. Lines 2-4 apply the taskloss, and lines 5-13 apply constraints loss and update the $\lambda_j$. The rate at which $\lambda_j$ is updated depends on \textit{PCT variable} ($\alpha$) decayed at each iteration's end.

\begin{algorithm}
\caption{PCT algorithm}
\label{alg:PCTalgorithm}
\begin{algorithmic}[1]
\For{each batch in data}
    \State Apply model on batch to get the logits
    \State Calculate Taskloss (CE/MSE/L1loss) 
    \State Backward propagate the loss
    \If {Not warm-up iteration}
        \State Get the next constraints batch
        \State Apply model on constraints batch
        \State $cl \gets 0$ \Comment{initialize constraints loss}
        \For {each constraint}
            \State $l \gets abs(q-p_1 \times p_2...\times p_n \times Pr)$
            \State $cl \gets cl + l \times \lambda_j$
            \State $\lambda_j \gets \alpha \times l$
        \EndFor
        \State Backward propagate the cl
    \EndIf
    \State Take optimizer step,and Reset gradients
    \State decay $\alpha$
\EndFor
\end{algorithmic}
\end{algorithm}

\subsection{Training Parameres}
\label{sec:appendix_training_paramteres}

\subsubsection{RuleTaker-pro}
To train RuleTaker-pro, we use RoBERTa Large for four epochs with a learning rate of $1e-5$. When we use PCT, the alpha (PCT variable) varies from 1.0 to 0.001 depending on the depth of the training dataset with higher depths training with smaller alphas.

\subsubsection{RuleBERT}
To train RuleBERT, we also use RoBERTa Large for four epochs, but we freeze the first 22 layers of the transformer. The learning rate varies between numbers $1e-6$ for higher depth datasets with more examples and $2e-6$ for lower depth datasets. When using PCT, the alpha is  0.01 for lower depths (1-3) and 0.001 for higher depths (4-5). In Table~\ref{tab:pd1}, the effect of alpha on the PCT Dev BA is shown. As shown, a higher alpha will help the model reach higher accuracy earlier. However, the best result is achieved with an alpha of 0.01.  

\begin{table*}[h]
\begin{center} 
\begin{tabular}{|c|c|c|c|c|c|c|c|c|c}
 \hline
 Depth3 & Epoch1 & Epoch2 & Epoch3 & Epoch4 & Epoch5 & Epoch6 \\ 
 \hline
 Baseline & 49 & 70 & 77.95 & 75.85 & 70.925 & 72.62 \\  
 PCT with $\alpha=0.1$ & 49 & \textbf{79.15} & 78.42 & 76.9 & 77 & 64.51  \\ 
 PCT with $\alpha=0.01$ & 49 & 79.32 & \textbf{80.87} & 79.32 & 78.17 & 78.57 \\  
 PCT with $\alpha=0.001$ & 49 & 70.90 & 78.55 & \textbf{80.85} & 78.55 & 78.75  \\ 
\hline

\end{tabular}

\end{center}

\caption{Accuracy obtained using PCT during training with different hyper-parameter ($\alpha$) for depth 3 of reasoning for 6 epochs on RuleBERT dataset. Normally we train our models for 4 epochs, but here we use 6 epochs to observe the learning process better.}
\label{tab:pd1}
\end{table*}

\subsection{Error Analysis Examples}
\label{sec:appendixerroranalysis}
\begin{figure}[h!]
    \includegraphics[width=0.45\textwidth]{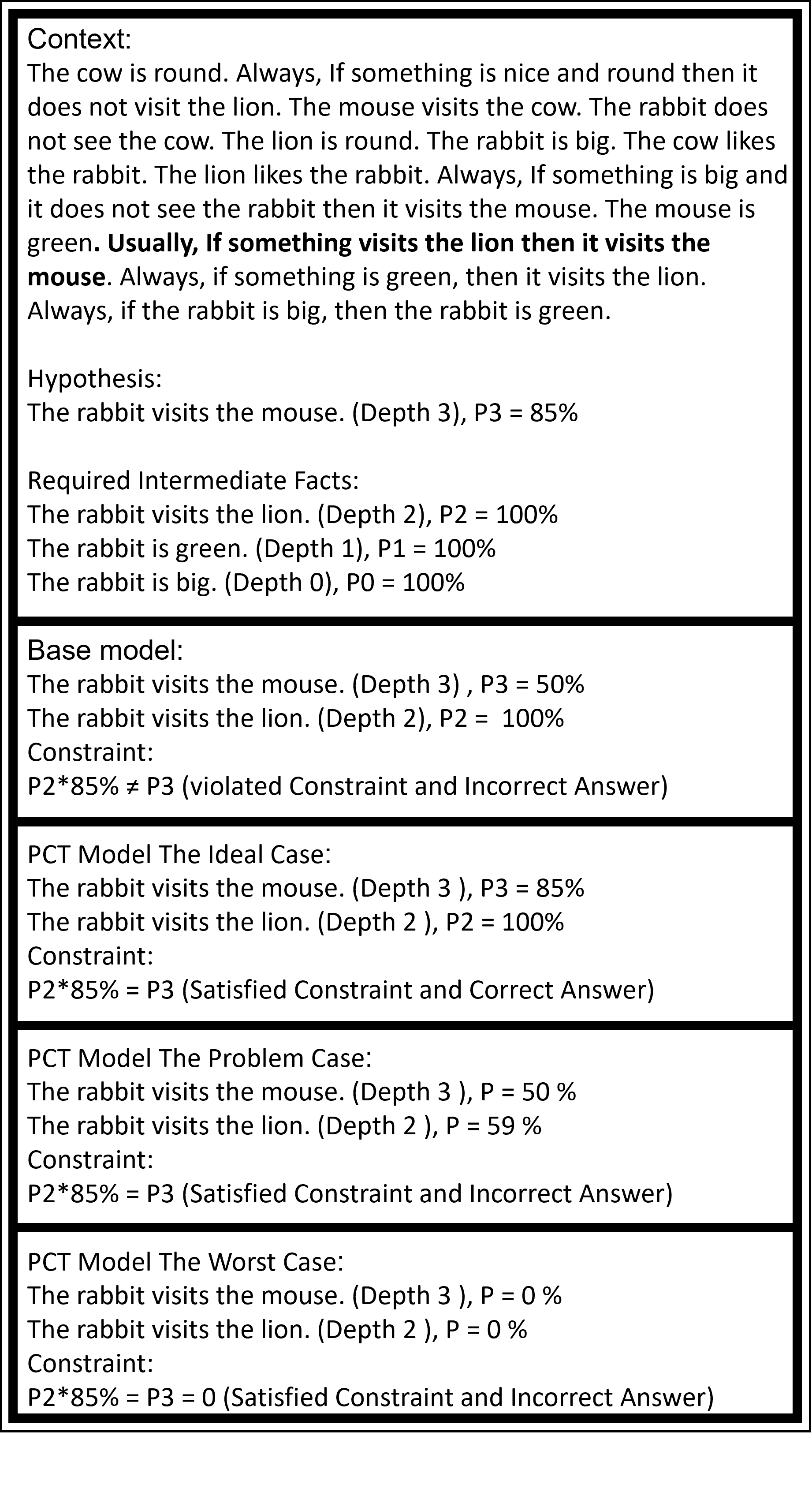}
    \caption{In the given example, the fact ``The rabbit visits the lion.'' can be inferred from the context with a probability of 1.00 at depth 2. Both the base model and the PCT model accurately predicted the probability of this fact. However, only the PCT model took into account the additional bold rule in the text, which led to an 0.85 probability for the hypothesis.}
    \label{fig:Error_analysis_example}
\end{figure}

We analyze an example shown in Figure~\ref{fig:Error_analysis_example} that benefited from PCT. Initially, the base model predicted 0.50 for the final answer, which was incorrect, as the answer should have been 0.85. After training the model using PCT, the model correctly predicted 0.85. This demonstrates the potential of the PCT model for incorporating additional constraints in the inference process. However, it should be noted that this is an ideal case that may not always be reproduced in practice. The PCT model can be adapted to alter the probability of the depth2 fact to satisfy the constraint if needed. In other scenarios, the model may keep the 0.50 prediction for depth 3 and change the prediction for depth 2. In this case, the model satisfies the constraint, yet the final prediction is incorrect. In the worst case, the model may predict 0.0 for all elements and still satisfy the constraint. 

It has been observed that the predicted probabilities of the PCT models are lower on average than those of the baseline models. This is due to the fact that lower predicted probabilities make it easier to satisfy the constraints, and thus, even models that improve overall accuracy tend to have lower average predicted probabilities.

\subsection{LLM Prompt Instructions and Additional Results}
\label{sec:appendixLLMRuleTakerpro}

To effectively evaluate LLMs like, we adjust our approach with our datasets to make them suitable for zero-shot and in-context settings for generative models. These adaptations involved adding a text explaining the task before the context. For RuleBERT, we use the following explanation, \textit{``Answer the following logical probabilistic question with only one word, True or False.''} and add the probability of the rules to their text. For RuleTaker-pro, we use \textit{``Answer the following logical probabilistic question in the format .\#\#, which is the probability of the question asked rounded to 2 decimals, for example, .13\%''}. After this text, we provide the context and pose the hypothesis as a question.

To test RuleBERT in LLMs, we included the probability of the rules in the text; Otherwise, the model has no way of extracting them. The results are shown in Table~\ref{tab:RuleBERTLLM_gpt3.5and4}.

\begin{table}[h]
\begin{center} \small
\resizebox{0.40\textwidth}{13mm}{
\begin{tabular}{|c|c|c|c|c|}
 \hline
Model &  GPT3.5 & GPT3.5*  & GPT4  \\ 
\hline
Depth1 & 19\% &  43\% & 29\%   \\ 
\hline
Depth2 & 58\% &  53\% & 46\%   \\ 
\hline
Depth3 & 58\% &  58\% & 60\%  \\ 
\hline
Depth4 & 51\% &  56\% & 46\%   \\ 
\hline
Depth5 & 56\% &  43\% & 58\%  \\ 
\hline
\end{tabular}
}
\end{center}
\caption{RuleBERT BA results are show for GPT3.5 and GPT4. * indicates few-shot setting.}
\label{tab:RuleBERTLLM_gpt3.5and4}
\end{table}

RuleTake-pro results for CA1 and CA10 are compared in Table~\ref{tab:LLMresults_ca1_ca10}. The only model that improves with regard to an increase in the threshold of the final answer is RoBERTa. This suggests that if the LLM can now predict the final answer, it would not predict anything close to it.

\begin{table}[h]
\centering

\begin{tabular}{|c|c|c|c|c|}
\hline
 & \multicolumn{4}{c|}{CA1}  \\
\hline
 & RoBERTa & GPT3.5 & GPT3.5* & GPT4  \\
\hline
D1 & \HeatCellGPT{44} & \HeatCellGPT{28} & \HeatCellGPT{41} & \HeatCellGPT{41}  \\
D2 & \HeatCellGPT{33} & \HeatCellGPT{20} & \HeatCellGPT{26} & \HeatCellGPT{27}  \\
D3 & \HeatCellGPT{28} & \HeatCellGPT{23} & \HeatCellGPT{25} & \HeatCellGPT{26} \\
D4 & \HeatCellGPT{27} & \HeatCellGPT{18} & \HeatCellGPT{20} & \HeatCellGPT{17}\\
D5 & \HeatCellGPT{28} & \HeatCellGPT{18} & \HeatCellGPT{20} & \HeatCellGPT{21}\\
\hline
& \multicolumn{4}{c|}{CA10} \\
\hline
D1 &  \HeatCellGPT{56} & \HeatCellGPT{33} & \HeatCellGPT{45} & \HeatCellGPT{43} \\
D2 &  \HeatCellGPT{52} & \HeatCellGPT{28} & \HeatCellGPT{36} & \HeatCellGPT{37} \\
D3 &  \HeatCellGPT{51} & \HeatCellGPT{25} & \HeatCellGPT{34} & \HeatCellGPT{34} \\
D4 &  \HeatCellGPT{52} & \HeatCellGPT{21} & \HeatCellGPT{33} &  \HeatCellGPT{29}\\
D5 &  \HeatCellGPT{53} & \HeatCellGPT{22} & \HeatCellGPT{31} &  \HeatCellGPT{29}\\
\hline

\end{tabular}

\caption{LLM results on RuleTaker-pro for CA1 and CA10 metrics. * indicates using few-shot examples. The chosen RoBERTa model is M5 trained with CE since it performs the best regarding CA1.}
\label{tab:LLMresults_ca1_ca10}
\end{table}

As we mentioned previously COT did not improve our models. Normally, one would expect CoT to always improve the results by adding a little extra reasoning and explanation for the LLM, but here it does not do the same for the following reasons: 1)The reasoning here is very complex and may require a combination of up to 5 rules and five facts, which are explained with long text to infer the final answer
2) In addition to the previous reason, the text of the rules is very large, and the context includes random rules mixed with useful rules that are not needed to answer the final question. 3)The task requires math and number extraction, which LLMs historically struggle with.

\cite{Shi2022StepGameAN} is another dataset who suffers from the first two reasons and as a results dose not improve with COT. Given these complexities, the provided instructions and examples in COT that we initially tried ended up actually hurting the model counterintuitively. For example, in some cases, the model would try to imitate the exact solutions in the few-shot examples to answer the questions and hallucinate the probabilities that don't exist in the text in its reasoning. Given these complexities, in these datasets, CoT is not just a baseline model and needs significant careful prompt engineering to be useful.

\subsection{Additional RuleTaker-pro Results}
\label{sec:appendixRuleTaker_detailed}
In Table~\ref{tab:ruletakerbaseBA}, The binary results for RuleTaker-pro trained with MSE and CE is shown.

\begin{table*}[h]
\begin{center}

\begin{tabular}{|c|c|c|c|c|c|c|c|c|c}
\hline
&\multicolumn{4}{c|}{CE Loss}&\multicolumn{4}{c|}{MSE Loss}\\
 \hline
 \hline
BA & M1 & M2 & M3  & Mmax  & M1 & M2 & M3  & Mmax  \\ 
\hline
Total & 76.93 &  82.65 & 88.74 & 91.05 & 76.19 &  84.84 & 87.73 & 91.39  \\ 
\hline
D1  & 97.19 &  94.85 & 92.18 & 93.39 & 97.28 &  95.92 & 92.64 & 94.33  \\ 
D2  & 75.58  & 89.11 & 91.26 & 91.26 & 74.41 &  90.91 & 91.88 & 91.74  \\ 
D3  & 68.19  & 77.35  & 89.42 & 91.00 & 42.88 &  81.93 & 88.59 & 90.34  \\
D4  & 65.16  & 71.35 & 84.93 & 88.70 & 38.04 &  74.38 & 81.82 & 89.17 \\ 
D5  & 58.61  & 65.05 & 80.96 & 88.31 & 57.70 &  66.96 & 76.43 & 88.21 \\ 
\hline
\hline
MSE & M1 & M2 & M3  & Mmax  & M1 & M2 & M3  & Mmax  \\ 
\hline
Total & 0.1574 &  0.1278 & 0.0965 & 0.0716 & 0.4693 &  0.6585 & 0.6298 & 0.0716  \\
\hline
D1  & 0.0866 &  0.0996 & 0.1076 & 0.0983 & 0.1992 &  0.0173 & 0.0190 & 0.0983  \\ 
D2  & 0.1939  & 0.1261 & 0.1065 & 0.0876 & 0.1902 &  0.0257 & 0.0247 & 0.0876  \\ 
D3  & 0.2352  & 0.1826 & 0.1149 & 0.0818 & 0.1915 &  0.0698 & 0.0313 & 0.0818  \\
D4  & 0.2511  & 0.2003 & 0.1281 & 0.0710 & 0.1910 &  0.0982 & 0.0423 & 0.0797 \\ 
D5  & 0.3082  & 0.2436 & 0.1428 & 0.710. & 0.1963 &  0.1237 & 0.0618 & 0.0710 \\ 
\hline
\hline
L1 & M1 & M2 & M3  & Mmax  & M1 & M2 & M3  & Mmax  \\ 
\hline
Total & 0.2505 &  0.2216 & 0.1903 & 0.1664 & 0.3628 &  0.1055 & 0.0798 & 0.1664  \\ 
\hline
D1  & 0.2004 &  0.2138 & 0.2236 & 0.2175 & 0.3693 &  0.0525 & 0.0581 & 0.2175  \\ 
D2  & 0.3118  & 0.2434 & 0.2243 & 0.2076 & 0.3638 &  0.0770 & 0.0786 & 0.2076  \\ 
D3  & 0.3528  & 0.3010 & 0.2316 & 0.1972 & 0.3642 &  0.1570 & 0.1032 & 0.1972  \\
D4  & 0.3672  & 0.3182 & 0.2443 & 0.1960 & 0.3659 &  0.2090 & 0.1326 & 0.1960 \\ 
D5  & 0.4136  & 0.3519 & 0.2495 & 0.1761 & 0.3737 &  0.2480 & 0.1627 & 0.1761 \\ 
\hline

\end{tabular}

\end{center}

\caption{The Binary accuracy, MSE and L1 of the baseline model trained and tested on the RuleTaker-pro dataset at different depths.}
\label{tab:ruletakerbaseBA}
\end{table*}

Additional detailed baseline and PCT results of RuleTaker-pro are shown in Tables~\ref{tab:ruletakerbase_full} and \ref{tab:ruletakerbase_PD_full}.

\begin{table*}[h]
\begin{center} 

\begin{tabular}{|c|c|c|c|c|c|c|c|c|c|c|}
\hline
 &\multicolumn{4}{c|}{CE (CA1)}&\multicolumn{4}{c|}{MSE (CA1)}\\
 \hline
D\slash M & M1 & M2 & M3  & Mmax  & M1 & M2 & M3  & Mmax   \\ 
\hline
Total & 38.21 &  \textbf{38.34} & 20.45 & 33.89 & 30.39 &  32.26 & 26.17 & 26.04 \\ 
\hline
D1  & \textbf{56.03} &  52.71 & 29.63 & 43.77 & 50.46 &  49.48 & 38.15 & 37.26 \\ 
D2  & 36.40  & \textbf{38.28} & 20.31 & 32.87 & 26.49 &  31.13 & 25.52 & 28.43  \\ 
D3  & 29.30  & \textbf{31.30}  & 14.98 & 28.39 & 18.81 &  22.06 & 18.98 & 19.90  \\
D4  & 27.49  & \textbf{28.53} & 14.03 & 27.11 & 18.54 &  21.37 & 17.79 & 17.32  \\ 
D5  & 24.97  & 26.78 & 14.70 & \textbf{28.29} & 19.83 &  21.14 & 19.33 & 15.50  \\ 
\hline
CS1  & 47.88  & 35.79  & 16.22 & 20.78 & 25.24 &  14.88& 14.47 & 12.97  \\ 
\hline
&\multicolumn{4}{c|}{CE (CA10)} &\multicolumn{4}{c|}{MSE (CA10)}\\
 \hline
D\slash M & M1 & M2 & M4  & Mmax  & M1 & M2 & M3  & Mmax   \\ 
\hline
Total & 46.45 &  49.69 & 49.95 & 53.25 & 58.15 &  62.75 & 66.67 & 74.80 \\ 
\hline
D1  & 61.56 &  59.55 & 53.41 & 56.17 & 91.76 &  84.45 & 80.94&\textbf{ 82.81}  \\ 
D2  & 45.76  & 52.08 & 52.42 & 51.59 & 53.60 &  69.97 & 77.25 & \textbf{77.32}  \\ 
D3  & 38.88  & 44.87  & 48.37 & 51.45 & 42.88 &  51.20 & 61.44 & \textbf{71.60}  \\
D4  & 37.75  & 42.45 & 47.36 & 51.97 & 38.04 &  46.51 & 51.50 & \textbf{69.39} \\ 
D5  & 33.63  & 38.67 & 43.80 & 53.07 & 32.62 &  37.05 & 43.30 & \textbf{63.64}\\ 
\hline
CS10  & 52.24  & 44.97 & 35.67 & 38.25 & 45.13 &  34.49 & 32.86 & 33.34  \\ 
\hline

\end{tabular}

\end{center}

\caption{The accuracy of the \textbf{baseline} models trained and tested on the RuleTaker-pro dataset. The rows show different test depths (depths 1 to 5). Total indicates the weighted average accuracy of all depths, and CS* shows the constraint satisfaction at the indicated thresholds. The best results for each depth are in bold.}
\label{tab:ruletakerbase_full}
\end{table*}

\begin{table*}[h]
\begin{center} 

\begin{tabular}{|c|c|c|c|c|c|c|c|c|c|c}
 \hline
 &\multicolumn{4}{c|}{CE+PCT (CA1)}&\multicolumn{4}{c|}{MSE+PCT (CA1)}\\
 \hline
D\slash M & M1 &  M2 & M3  & Mmax  &  M1 & M2 & M3  & Mmax  \\ 
\hline
Total & 38.0 &39.5 &  41.1 &  37.6 &  37.4 &34.7 &  36.4 &  34.3\\ 
 \hline
D1 & 53.3 & 50.8 &  50.5 &  46.9 & 56.50 &49.8 &  52.6 &  37.6\\ 
D2 & 37.4 &40.4 &  42.2 &  37.0 &  35.99 &34.2 &  38.1 &  33.8\\ 
D3 & 26.4 &32.9 &  36.0 &  32.4& 25.9 &25.8 &  26.5 &  32.6\\ 
D4 & 26.5 &31.9 &  33.9 &  31.8 &24.1 &25.5 &  24.9 &  31.6 \\ 
D5 & 23.3 &30.4 &  33.4 &  31.4 & 22.0 &24.0&  24.0 &  33.1\\ 
 \hline
CS1  & 44.9  & 42.6 & 34.5 & 35.2 & 20.5&  19.3 & 15.4 & 13.0 \\ 

\hline
 &\multicolumn{4}{c|}{CE+PCT (CA10)}&\multicolumn{4}{c|}{MSE+PCT (CA10)}\\
 \hline
D\slash M & M1 &  M2 & M3  & Mmax  &  M1 & M2 & M3  & Mmax  \\ 
\hline
Total &46.6  &  50.8 &  52.5 &  52.9 &58.9 &  63.3 &  67.2 &  68.7 \\ 
\hline
D1 & 59.7 &  57.8 &  50.5 & 57.9 & 92.4&  82.7 &  83.5 &  70.9 \\ 
D2 & 47.78 &  51.4 &  42.2 &  51.8 & 57.7 &  73.2 &  76.1 &  68.2 \\ 
D3 & 39.2 &  47.4&  50.5 &  50.0 & 41.5 &  52.2 &  60.4 &  68.6 \\ 
D4 & 36.2 &  47.0 &  50.1 &  50.4 & 36.1 &  47.3&  51.4&  70.0 \\ 
D5 & 35.1 &  47.0&  48.6&  49.8 &34.0 &  38.1 &  44.6 &  63.8 \\ 
\hline
CS10 & 49.7 & 47.3 & 45.6& 46.8 & 49.6 &  36.0 & 34.9 & 33.7 \\ 
\hline

\end{tabular}

\end{center}

\caption{RuleTaker-pro results trained with \textbf{PCT}. The rows show different test depths (depths 1 to 5). Total indicates the weighted average accuracy of all depths, and CS* shows the constraint satisfaction at the indicated thresholds. } 
\label{tab:ruletakerbase_PD_full}
\end{table*}

\end{document}